\documentclass[10pt,twocolumn,letterpaper]{article}

\usepackage[pagenumbers]{cvpr}      

\usepackage{tabularx}
\usepackage{tcolorbox}

\usepackage[table]{xcolor}
\usepackage{algorithm}
\usepackage{algorithmic}
\usepackage{multirow}
\definecolor{cvprblue}{rgb}{0.21,0.49,0.74}
\usepackage[pagebackref,breaklinks,colorlinks,allcolors=cvprblue]{hyperref}

\def\confName{CVPR}
\def\confYear{2027}

\title{Think More or Look Again?\\
Training-Free Test-Time Scaling for Multi-Image Medical Reasoning}

\author{
Ali Khoramfar$^{1}$ \quad
Mohammad Mahdi Salmani-Zarchi$^{1}$ \\[2pt]
Mohammad Javad Dousti$^{1}$ \quad
Alireza Mohamadian$^{2}$ \quad
Heshaam Faili$^{1}$ \\[5pt]
{\normalsize $^{1}$Department of Electrical and Computer Engineering, College of Engineering,}\\
{\normalsize University of Tehran, Tehran, Iran}\\[2pt]
{\normalsize $^{2}$Advanced Diagnostic and Interventional Radiology Research Center (ADIR),}\\
{\normalsize Tehran University of Medical Sciences, Tehran, Iran}\\[3pt]
{\tt\small $^{1}$\{khoramfar, m.salmani78, mjdousti, hfaili\}@ut.ac.ir, $^{2}$alirezamohamadian.md@gmail.com}
}

\begin{document}
\maketitle
\begin{abstract}
Medical image interpretation often requires integrating complementary evidence distributed across multiple imaging sequences. Although test-time scaling may improve model performance by allocating additional inference computation, extending textual reasoning alone can reduce visual grounding. We study staged visual inference, a training-free strategy in which a vision-language model first interprets one imaging sequence and subsequently receives complementary images while preserving its inference prefix. Experiments with Qwen3.5-9B and InternVL3.5-8B on a balanced cohort of 90 patients with brain MRI examinations show that the timing and direction of evidence presentation materially affect diagnostic performance. Presenting T2 before T1CE improves balanced accuracy over the corresponding joint-input baseline by 16.7 and 11.1 percentage points for Qwen3.5-9B and InternVL3.5-8B, respectively. These gains are not reproduced by text-only continuation, which consistently increases prediction instability and reduces accuracy. Built-in thinking also incurs substantially greater token and latency costs and can degrade staged inference. Evaluation on prostate and liver MRI demonstrates that the benefits transfer selectively across datasets and models. Overall, staged visual presentation can support diagnostic revision without retraining or architectural modification, but its effectiveness depends on sequence order, model behavior, and the clinical task. 
\end{abstract}

\section{Introduction}

\label{sec:intro}
\begin{figure}[t]
    \centering
    \includegraphics[trim=0 0.2cm 0 0,width=\columnwidth]{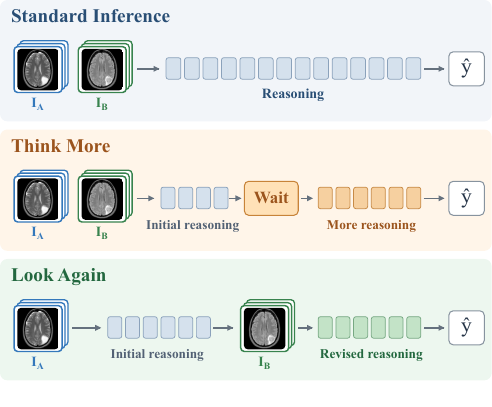}
    \caption{Standard inference uses joint inputs; \emph{Think More} extends textual reasoning, while \emph{Look Again} adds complementary images mid-inference.}
    \label{fig:thinkmore_lookagain}
\end{figure}

\begin{figure*}[t]
    \centering
    \includegraphics[trim=0 0 0 0,width=0.95\textwidth]{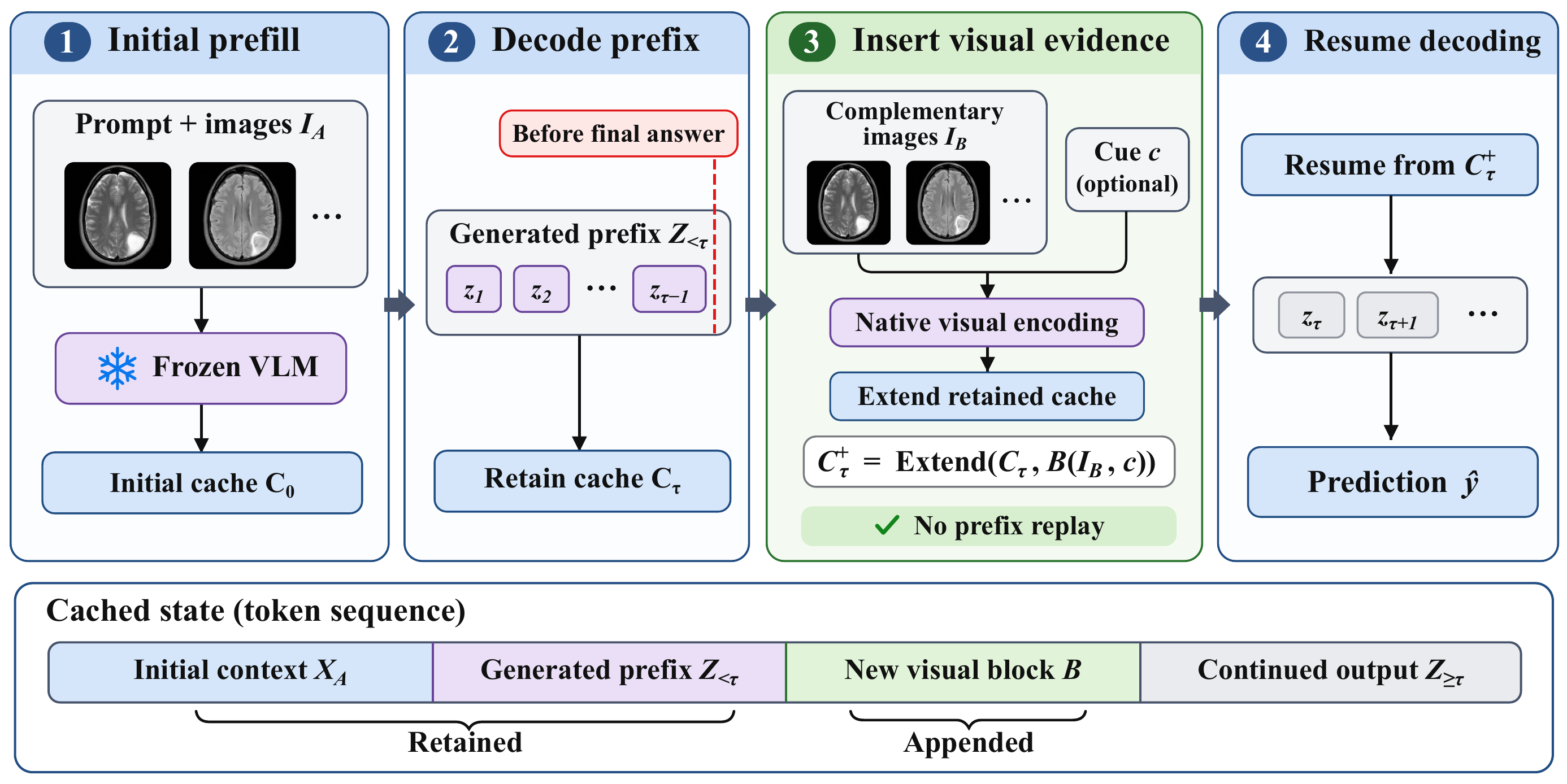}
    \caption{After initial prefill and prefix decoding, complementary images (and an optional cue) are appended to the retained cache; decoding resumes without prefix replay.}
    \label{fig:injection}
\end{figure*}

Test-time scaling has emerged as a promising paradigm for improving model performance by allocating additional computation during inference~\cite{snell2024scalingllmtesttimecompute}. Rather than committing immediately to a prediction, a model can use this additional budget to extend its reasoning, explore alternative solution paths, or revisit intermediate conclusions~\cite{muennighoff2025s1}. This shifts part of the scaling problem from training increasingly specialized models to determining how the capabilities of an existing model can be used more effectively at test time.

For Vision-Language Models (VLMs), making better use of existing capabilities involves more than extending the reasoning process. These models form their predictions by combining language-based reasoning with information extracted from the visual input~\cite{zhang_vision-language_2024,yin_survey_2024}. Additional test-time computation therefore creates a choice: Should the model \emph{think more} about its current interpretation, or \emph{look again} at the visual evidence before reaching a final prediction? This question is particularly important in the medical domain, where diagnostic decisions depend heavily on information contained in the images.

Medical imaging makes this choice especially consequential because diagnostic evidence is rarely confined to a single image. A clinical examination may contain multiple views, slices, or imaging sequences, each contributing different information about the same case~\cite{hartsock_vision-language_2024}. The challenge is therefore not only to make this evidence available to the model, but also to determine how it should be incorporated as the diagnostic interpretation develops.

Existing approaches commonly address these challenges through domain-specific training, specialized volumetric representations, or learned mechanisms for selecting diagnostically relevant images~\cite{bai2024m3d,xin2025med3dvlm,chen2025pmbb}. While these directions have advanced medical image understanding, they typically require additional data, optimization, or task-specific components to integrate information across multiple images and complementary imaging sequences. Test-time scaling offers a different possibility: Can a frozen VLM reason more effectively over such complex imaging studies without additional training or architectural modification?

Real-world medical imaging further complicates this question. Relevant findings may be subtle, distributed across an examination, or visible only through complementary visual information. Effective test-time scaling must therefore do more than prolong the model's reasoning: It must determine how visual evidence should enter and inform the evolving diagnostic interpretation.

In this work, we investigate this possibility using a curated cohort of multi-sequence brain magnetic resonance imaging (MRI) examinations. We compare three inference settings: joint-input standard inference, extended textual continuation (\emph{Think More}), and staged introduction of complementary visual evidence (\emph{Look Again}) (\Cref{fig:thinkmore_lookagain}).

Our findings show that evidence timing matters: \emph{Look Again} can substantially outperform standard joint-input inference, whereas \emph{Think More} fails to reproduce these gains. Our main contributions are summarized as follows: \vspace{0.1cm} 
\begin{itemize} 
\setlength\itemsep{0.1cm}

\item \textbf{Staged Visual Inference:} We propose a training-free inference strategy that presents complementary imaging sequences in stages, enabling newly introduced visual evidence to refine an existing diagnostic interpretation.
\item \textbf{Mechanistic Characterization:} We show that the benefits of staging arise from visually grounded diagnostic revision rather than additional generation steps: Text-only continuation destabilizes predictions, while prolonged built-in thinking can dilute visual attention and increase computational cost.
\item \textbf{Multi-task Evaluation:} We evaluate two open-source VLMs on multi-sequence brain, prostate, and liver MRI cohorts, identifying substantial improvements in favorable configurations alongside important dependencies on sequence order, model architecture, and clinical task. \end{itemize}

\section{Related Work}
\label{sec:related}

\paragraph{Test-time scaling and visual grounding.}
Test-time scaling improves large language model (LLM) reasoning by allocating additional inference computation, such as extending generation with continuation cues (e.g., budget forcing in s1~\cite{muennighoff2025s1}) or exploring alternative reasoning trajectories~\cite{snell2024scalingllmtesttimecompute}. When transferred to VLMs, however, prolonged textual reasoning often shifts attention away from image tokens and degrades visual grounding~\cite{sammani2026scaling,bi2026lookback}. In medical benchmarks, extended reasoning budgets and purely textual budget forcing similarly exhibit diminishing returns or overthinking~\cite{oh2026medicaltts}. Rather than scaling computation through longer textual reasoning alone, our work evaluates whether introducing a complementary imaging sequence mid-inference enables more effective diagnostic revision than text-only continuation or built-in thinking mode.

\paragraph{Iterative visual reasoning and multi-image order sensitivity.}
Several recent methods enhance multimodal reasoning through iterative perception, such as interleaving attention-selected crops with rationales~\cite{gao2025icot}, dynamically zooming into informative regions~\cite{luo2025visualtts,lin2026adaptvision}, decoupling visual search from reasoning~\cite{avogaro2026sparc}, or revisiting tokens and perception traces within a given image~\cite{ghosal2026visref,jiang2026ttsp}. These approaches operate on visual content that is already present in the input and focus on spatial refinement. In clinical imaging, however, diagnostic findings are frequently distributed across distinct, complementary sequences (e.g., T2-weighted and post-contrast T1 MRI) rather than localized within a single high-resolution view. Moreover, multi-image VLMs are sensitive to input ordering and exhibit pronounced position bias across images~\cite{tian2025positionbias}. To separate the effect of evidence timing from presentation order, our primary brain MRI evaluation tests both staging directions against order-matched joint baselines.

\paragraph{Multi-sequence medical imaging with VLMs.} In clinical radiology, accurate diagnosis on volumetric computed tomography (CT) and multi-sequence MRI requires synthesizing complementary findings across distinct acquisitions and hundreds of slices. Prior medical VLM research has largely tackled this complexity through specialized engineering: training domain-adapted volumetric or multi-sequence models~\cite{hartsock_vision-language_2024,bai2024m3d,xin2025med3dvlm,lteif2026remind}, developing learned slice-selection modules to filter relevant frames~\cite{chen2025pmbb,gloeckler2026xglass}, or benchmarking diagnostic accuracy on multi-sequence cohorts~\cite{elboardy2025benchmarking}. While effective within their target settings, such pipelines depend on costly data curation, extensive retraining, or task-specific architectural components that limit plug-and-play deployment across clinical protocols. At the same time, modern general-purpose VLMs natively accept long multi-image contexts, raising a natural yet underexplored question: To what extent can their existing capabilities be harnessed for real-world multi-sequence diagnosis without parameter updates or learned pre-selectors? Rather than building specialized medical architectures, we frame inference-time sequence organization as a zero-training probe of frozen VLMs---evaluating both the potential and the practical boundary conditions of visual presentation across clinical multi-sequence tasks.

\begin{table*}[t]
  \caption{Primary comparison across single-sequence, joint, and staged inference on the brain MRI cohort. Following \Cref{eq:inference_conditions}, underlined pairs denote joint presentation and arrows ($\rightarrow$) denote staged injection of the complementary sequence after an initial interpretation. Best Balanced Accuracy and Macro-F1 per model are in \textbf{bold}.}
  \label{tab:primary_comparison}
  \centering
  \small
  \setlength{\tabcolsep}{5pt}
  \renewcommand{\arraystretch}{1.12}
  \begin{tabularx}{\textwidth}{ll l *{5}{>{\centering\arraybackslash}X}}
    \toprule
    & & & \multicolumn{2}{c}{\textbf{Overall (\%)}} & \multicolumn{2}{c}{\textbf{Per-Class Recall (\%)}} & \\
    \cmidrule(lr){4-5} \cmidrule(lr){6-7}
    \textbf{Model} & \textbf{Condition} & \textbf{Input Flow} &
    \textbf{\begin{tabular}{@{}c@{}}Balanced\\Accuracy\end{tabular}} &
    \textbf{\begin{tabular}{@{}c@{}}Macro-F1\end{tabular}} &
    \textbf{GBM} & \textbf{MET} &
    \textbf{\begin{tabular}{@{}c@{}}Median\\Latency (s)\end{tabular}} \\
    \midrule
    \multirow{6}{*}{Qwen3.5-9B}
      & \multirow{2}{*}{Single}
        & $\text{T1CE}$                            & 64.4 & 60.0 & 31.1 & 97.8 & 11.7 \\
      & & $\text{T2}$                              & 65.6 & 65.5 & 62.2 & 68.9 & 12.4 \\
      \cmidrule{2-8}
      & \multirow{2}{*}{Joint}
        & ($\underline{\text{T1CE}, \text{T2}})$   & 64.4 & 60.0 & 31.1 & 97.8 & 12.7 \\
      & & ($\underline{\text{T2}, \text{T1CE}})$   & 58.9 & 50.5 & 17.8 & 100.0 & 12.1 \\
      \cmidrule{2-8}
      & \cellcolor{blue!6} & \cellcolor{blue!6}$\text{T1CE} \rightarrow \text{T2}$ & \cellcolor{blue!6}64.4 & \cellcolor{blue!6}63.0 & \cellcolor{blue!6}44.4 & \cellcolor{blue!6}84.4 & \cellcolor{blue!6}21.8 \\
      & \cellcolor{blue!6}\multirow{-2}{*}{\textbf{Staged (Ours)}} & \cellcolor{blue!6}$\text{T2} \rightarrow \text{T1CE}$ & \cellcolor{blue!6}\textbf{75.6} & \cellcolor{blue!6}\textbf{75.5} & \cellcolor{blue!6}71.1 & \cellcolor{blue!6}80.0 & \cellcolor{blue!6}23.9 \\
    \midrule
    \multirow{6}{*}{InternVL3.5-8B}
      & \multirow{2}{*}{Single}
        & $\text{T1CE}$                            & 55.6 & 45.9 & 13.3 & 97.8 & 4.8 \\
      & & $\text{T2}$                              & 57.8 & 56.0 & 37.8 & 77.8 & 4.9 \\
      \cmidrule{2-8}
      & \multirow{2}{*}{Joint}
        & ($\underline{\text{T1CE}, \text{T2}})$   & 60.0 & 54.2 & 24.4 & 95.6 & 6.7 \\
      & & ($\underline{\text{T2}, \text{T1CE}})$   & 54.4 & 48.3 & 20.0 & 88.9 & 6.1 \\
      \cmidrule{2-8}
      & \cellcolor{blue!6} & \cellcolor{blue!6}$\text{T1CE} \rightarrow \text{T2}$ & \cellcolor{blue!6}\textbf{65.6} & \cellcolor{blue!6}\textbf{65.5} & \cellcolor{blue!6}62.2 & \cellcolor{blue!6}68.9 & \cellcolor{blue!6}11.9 \\
      & \cellcolor{blue!6}\multirow{-2}{*}{\textbf{Staged (Ours)}} & \cellcolor{blue!6}$\text{T2} \rightarrow \text{T1CE}$ & \cellcolor{blue!6}\textbf{65.6} & \cellcolor{blue!6}65.2 & \cellcolor{blue!6}55.6 & \cellcolor{blue!6}75.6 & \cellcolor{blue!6}11.8 \\
    \bottomrule
  \end{tabularx}
\end{table*}

\section{Methodology}
\label{sec:method}

\subsection{Formulation and Inference Conditions}
\label{sec:formulation}

Let $I_A$ and $I_B$ denote two ordered imaging sequences (or sequence groups) that provide complementary evidence for the same case, and let $\hat{y}$ denote the final case-level prediction. Using a fixed VLM, we compare three ways of presenting the available evidence.

In the \emph{single-sequence} condition, the model receives either $I_A$ or $I_B$. In the \emph{joint} condition, both sequences are available from the beginning. In the \emph{staged} condition, the model first examines $I_A$ and reaches an intermediate inference state $z_A$. Before the model commits to its final prediction, the complementary sequence $I_B$ is introduced and inference continues toward $\hat{y}$.

The three conditions are summarized as
\begin{equation}
\begin{aligned}
\emph{Single:}\quad & I_A \rightarrow \hat{y},\\
\emph{Joint:}\quad & (I_A,I_B) \rightarrow \hat{y},\\
\emph{Staged:}\quad & I_A \rightarrow z_A \rightarrow I_B \rightarrow \hat{y}.
\end{aligned}
\label{eq:inference_conditions}
\end{equation}

The staged condition retains the inference developed from $I_A$, allowing the newly introduced evidence to refine an existing interpretation rather than initiating a new prediction process. In our implementation, $I_B$ is introduced either directly or with a short continuation cue (such as \texttt{Wait}), and inference resumes from the retained state. This creates a visual form of test-time scaling in which additional computation processes new evidence instead of only extending the textual trajectory.

\subsection{Cache-Preserving Visual Injection}
\label{sec:visual_injection}

\Cref{fig:injection} details the staged execution. Initial prefill processes the diagnostic prompt and images $I_A$, forming context $X_A$ and an initial inference cache $C_0$. The model then decodes a textual prefix $Z_{<\tau}$, whose generated content and retained state instantiate $z_A$. At the intervention point $\tau$---triggered when decoding reaches the commitment boundary (\texttt{Final Answer:} in standard mode or \texttt{</think>} in thinking mode)---generation pauses before the boundary is committed and the current cache $C_\tau$ is retained.

Complementary images $I_B$ are processed through the model's native visual encoder and packaged with an optional continuation cue $c$ into an appended block $B(I_B,c)$. The cache update is
\begin{equation}
    C_\tau^{+} = \operatorname{Extend}\!\left(C_\tau, B(I_B,c)\right),
    \label{eq:cache_extension}
\end{equation}
where $\operatorname{Extend}$ executes a single incremental forward pass over $B(I_B,c)$ conditioned on $C_\tau$. Across architectures, this updates multimodal states and positional coordinates without prefix replay---advancing hybrid recurrent Gated DeltaNet states alongside full-attention key--value (KV) caches and 3D multimodal rotary position embedding (M-RoPE) offsets in Qwen3.5-9B~\cite{qwen2026qwen35}, or injecting Vision Transformer (ViT) embeddings along contiguous 1D cache positions in InternVL3.5-8B~\cite{opengvlab2025internvl35} (\Cref{sec:supp_decoding}). Decoding resumes from $C_\tau^{+}$ to produce $Z_{\ge\tau}$ and the final prediction $\hat{y}$. Text-only continuation appends a cue without complementary images, while cue-free staging appends the visual block without $c$. All conditions use the frozen model's existing visual and language components.

\subsection{Datasets and Clinical Tasks}
\label{sec:datasets}

\paragraph{Primary brain MRI cohort.}
Our primary evaluation uses a curated multi-center cohort of 90 patients, balanced between 45 glioblastoma (GBM) and 45 brain metastasis (MET) cases. Each patient includes paired T1-weighted post-contrast (T1CE) and T2-weighted (T2) MRI volumes, and all diagnostic labels are histopathologically confirmed from surgical resection or biopsy. This independently collected clinical cohort provides a controlled primary test set with a reduced risk of data contamination.

\paragraph{External evaluation.}
We additionally evaluate the staged presentation protocol on two public multi-sequence MRI datasets. OpenSwissHCC contains 132 liver MRI examinations for hepatocellular carcinoma (HCC) classification, including 63 HCC-positive and 69 HCC-negative cases~\cite{tomography12090133}; we use the complementary arterial-phase (Art.) and portal-venous-phase (PV) T1-weighted volumes. Prostate Imaging: Cancer AI (PI-CAI) is a large multi-center prostate MRI resource for detecting clinically significant prostate cancer (csPCa)~\cite{saha2023artificial}; we evaluate the official \texttt{fold~0} partition of 300 cases (87 positive and 213 negative) using the complementary T2-weighted (T2W), apparent diffusion coefficient (ADC), and high-$b$-value (HBV) volumes for binary case-level classification. Across all three cohorts, all available slices from the selected sequences are used without subsampling, and no lesion annotations, masks, or clinical metadata are provided to the model. Extended specifications and preprocessing details are provided in \Cref{sec:supp_datasets}.

\begin{table*}[t]
  \caption{Textual continuation versus staged visual presentation on brain MRI. Underlined pairs denote joint inputs. Text-only continuation reduces accuracy relative to the corresponding single or joint baseline, whereas T2-to-T1CE staging improves performance even without a cue (\textemdash). Best results per model are in \textbf{bold}.}
  \label{tab:text_vs_staged}
  \centering
  \small
  \setlength{\tabcolsep}{4.5pt}
  \renewcommand{\arraystretch}{1.14}
  \begin{tabularx}{\textwidth}{ll cc *{6}{>{\centering\arraybackslash}X}}
    \toprule
    & & & & \multicolumn{3}{c}{\textbf{Qwen3.5-9B}} & \multicolumn{3}{c}{\textbf{InternVL3.5-8B}} \\
    \cmidrule(lr){5-7} \cmidrule(lr){8-10}
    \textbf{Condition} & \textbf{Input Flow} & \textbf{Cue} & \textbf{New Img.} &
    \textbf{Bal. Acc.} & \textbf{Macro-F1} & \textbf{Lat. (s)} &
    \textbf{Bal. Acc.} & \textbf{Macro-F1} & \textbf{Lat. (s)} \\
    \midrule
    \multirow{4}{*}{Text-Only}
      & $\text{T1CE}$                         & \texttt{Wait} & --- & 48.9 & 48.0 & 25.1 & 48.9 & 47.6 & 7.6 \\
      & $\text{T2}$                           & \texttt{Wait} & --- & 40.0 & 39.5 & 24.4 & 46.7 & 46.6 & 7.9 \\
      & ($\underline{\text{T1CE}, \text{T2}})$ & \texttt{Wait} & --- & 43.3 & 39.4 & 27.0 & 53.3 & 52.5 & 10.5 \\
      & ($\underline{\text{T2}, \text{T1CE}})$ & \texttt{Wait} & --- & 50.0 & 45.8 & 31.3 & 48.9 & 48.9 & 9.3 \\
    \midrule
    \cellcolor{blue!6} & \cellcolor{blue!6}$\text{T1CE} \rightarrow \text{T2}$ & \cellcolor{blue!6}---           & \cellcolor{blue!6}$\checkmark$ & \cellcolor{blue!6}60.0 & \cellcolor{blue!6}55.7 & \cellcolor{blue!6}20.6 & \cellcolor{blue!6}64.4 & \cellcolor{blue!6}64.4 & \cellcolor{blue!6}11.7 \\
    \cellcolor{blue!6} & \cellcolor{blue!6}$\text{T1CE} \rightarrow \text{T2}$ & \cellcolor{blue!6}\texttt{Wait} & \cellcolor{blue!6}$\checkmark$ & \cellcolor{blue!6}64.4 & \cellcolor{blue!6}63.0 & \cellcolor{blue!6}21.8 & \cellcolor{blue!6}65.6 & \cellcolor{blue!6}65.5 & \cellcolor{blue!6}11.9 \\
    \cellcolor{blue!6} & \cellcolor{blue!6}$\text{T2} \rightarrow \text{T1CE}$ & \cellcolor{blue!6}---           & \cellcolor{blue!6}$\checkmark$ & \cellcolor{blue!6}\textbf{76.7} & \cellcolor{blue!6}\textbf{76.7} & \cellcolor{blue!6}21.8 & \cellcolor{blue!6}\textbf{67.8} & \cellcolor{blue!6}\textbf{67.3} & \cellcolor{blue!6}12.1 \\
    \cellcolor{blue!6}\multirow{-4}{*}{\textbf{Staged (Ours)}}
      & \cellcolor{blue!6}$\text{T2} \rightarrow \text{T1CE}$ & \cellcolor{blue!6}\texttt{Wait} & \cellcolor{blue!6}$\checkmark$ & \cellcolor{blue!6}75.6 & \cellcolor{blue!6}75.5 & \cellcolor{blue!6}23.9 & \cellcolor{blue!6}65.6 & \cellcolor{blue!6}65.2 & \cellcolor{blue!6}11.8 \\
    \bottomrule
  \end{tabularx}
\end{table*}

\begin{table*}[t]
  \caption{Paired prediction changes relative to order-matched joint baselines. Underlined pairs denote joint presentation as a single bundle. Best values among staged conditions per model are in \textbf{bold}. $\text{W}\rightarrow\text{C}$ and $\text{C}\rightarrow\text{W}$ denote wrong-to-correct and correct-to-wrong transitions.}
  \label{tab:paired_vs_joint}
  \centering
  \small
  \setlength{\tabcolsep}{5pt}
  \renewcommand{\arraystretch}{1.18}
  \begin{tabularx}{\textwidth}{l l c c >{\centering\arraybackslash}X c c >{\centering\arraybackslash}X}
    \toprule
    & & \multicolumn{3}{c}{\textbf{Qwen3.5-9B}} & \multicolumn{3}{c}{\textbf{InternVL3.5-8B}} \\
    \cmidrule(lr){3-5} \cmidrule(lr){6-8}
    \textbf{Condition} & \textbf{Comparison vs. Joint} &
    \textbf{W$\rightarrow$C\,$\uparrow$} & \textbf{C$\rightarrow$W\,$\downarrow$} & \textbf{Net ($\Delta$Acc.)} &
    \textbf{W$\rightarrow$C\,$\uparrow$} & \textbf{C$\rightarrow$W\,$\downarrow$} & \textbf{Net ($\Delta$Acc.)} \\
    \midrule
    \multirow{2}{*}{Text-Only}
      & ($\underline{\text{T1CE}, \text{T2}}) + \texttt{Wait}$ vs.\ ($\underline{\text{T1CE}, \text{T2}})$ & 23 & 42 & $-$19 ($-$21.1\%) & 19 & 25 & $-$6 ($-$6.7\%) \\
      & ($\underline{\text{T2}, \text{T1CE}}) + \texttt{Wait}$ vs. ($\underline{\text{T2}, \text{T1CE}})$ & 30 & 38 &  $-$8 ($-$8.9\%)  & 26 & 31 & $-$5 ($-$5.6\%) \\
    \midrule
    \cellcolor{blue!6} & \cellcolor{blue!6}$\text{T1CE} \rightarrow \text{T2}$ vs. ($\underline{\text{T1CE}, \text{T2}})$ & \cellcolor{blue!6}12 & \cellcolor{blue!6}12 & \cellcolor{blue!6}0 (+0.0\%) & \cellcolor{blue!6}\textbf{19} & \cellcolor{blue!6}14 & \cellcolor{blue!6}+5 (+5.6\%) \\
    \cellcolor{blue!6}\multirow{-2}{*}{\textbf{Staged (Ours)}}
      & \cellcolor{blue!6}$\text{T2} \rightarrow \text{T1CE}$ vs. ($\underline{\text{T2}, \text{T1CE}})$ & \cellcolor{blue!6}\textbf{25} & \cellcolor{blue!6}\textbf{10} & \cellcolor{blue!6}\textbf{+15 (+16.7\%)} & \cellcolor{blue!6}\textbf{19} & \cellcolor{blue!6}\textbf{9} & \cellcolor{blue!6}\textbf{+10 (+11.1\%)} \\
    \bottomrule
  \end{tabularx}
\end{table*}

\subsection{Models and Inference Protocol}
\label{sec:models_protocol}

We evaluate Qwen3.5-9B~\cite{qwen2026qwen35} and InternVL3.5-8B~\cite{opengvlab2025internvl35}, two open-source VLMs with native support for multi-image inputs. All evaluations are performed without parameter updates.

Each sequence is preceded by its sequence name, with its slices presented as independent frames in anatomical order. The diagnostic prompt, candidate labels, output format, and decoding configuration remain fixed across the compared conditions. Primary experiments use deterministic greedy decoding with thinking mode disabled and a fixed random seed across conditions to ensure reproducibility.

Image preprocessing is fixed before inference and does not depend on the ground-truth label or model output. Thinking-mode ablations follow the corresponding model-specific protocol and are reported separately from the primary thinking-disabled comparisons. Full preprocessing parameters, prompts, and implementation details are included in \Cref{sec:supp_preprocessing} and \Cref{sec:supp_protocol}.

\subsection{Evaluation Conditions}
\label{sec:conditions}

The primary comparison includes three conditions: \emph{single-sequence}, \emph{joint}, and \emph{staged} presentation. Single-sequence inference uses each available sequence independently. Joint inference provides all complementary sequences from the beginning. Staged inference presents one sequence first and introduces the complementary sequence before the final prediction. Unless otherwise specified, staged runs use the minimal \texttt{Wait} continuation cue as the default setting.

For the primary brain MRI cohort, we evaluate both T1CE-to-T2 and T2-to-T1CE directions, together with the corresponding joint input orders. Additional analyses examine textual continuation without new images, alternative continuation cues, removal of the cue, and the model's built-in thinking mode. These analyses are reported separately from the primary comparison. Extended ablations, sensitivity analyses, and per-case transition results are provided in \Cref{sec:supp_ablations} and \Cref{sec:supp_stats}.

\subsection{Metrics and Statistical Analysis}
\label{sec:metrics}

The patient is the unit of evaluation. Our primary diagnostic metric is balanced accuracy,
\begin{equation}
\mathrm{BA}
=
\frac{1}{C}
\sum_{c=1}^{C}
\mathrm{Recall}_c,
\label{eq:balanced_accuracy}
\end{equation}
where $C$ is the number of diagnostic classes. Balanced accuracy assigns equal weight to each class and is therefore comparable across binary, imbalanced, and multiclass tasks. We additionally report macro-F1,
\begin{equation}
\mathrm{MacroF1}
=
\frac{1}{C}
\sum_{c=1}^{C}
F1_c.
\label{eq:macro_f1}
\end{equation}
Macro-F1 complements balanced accuracy by incorporating both precision and recall while weighting all classes equally.

Because all inference conditions are evaluated on the same cases, we also report paired prediction changes, including wrong-to-correct ($\text{W}\rightarrow\text{C}$) and correct-to-wrong ($\text{C}\rightarrow\text{W}$) transitions.

We report recorded median inference latency within each model. Macro-F1 is recomputed from case-level predictions, assigning zero F1 to supported classes with no predicted positives. Additional uncertainty estimates and statistical analyses, including 95\% confidence intervals for the primary comparisons, are provided in \Cref{sec:supp_stats}.

\section{Results}
\label{sec:results}

\begin{table}[t]
  \caption{Effect of built-in thinking (\emph{Think ON} vs.\ \emph{OFF}) along the $\text{T2}$-first dual-sequence pathway. Best Balanced Accuracy and Macro-F1 per model are in \textbf{bold}.}
  \label{tab:thinking_ablation}
  \centering
  \small
  \setlength{\tabcolsep}{3.5pt}
  \renewcommand{\arraystretch}{1.02}
  \begin{tabularx}{\columnwidth}{l c *{4}{>{\centering\arraybackslash}X}}
    \toprule
    \textbf{Input Flow} & \textbf{Think} &
    \textbf{\begin{tabular}{@{}c@{}}Bal.\\Acc.\end{tabular}} &
    \textbf{\begin{tabular}{@{}c@{}}Macro-\\F1\end{tabular}} &
    \textbf{\begin{tabular}{@{}c@{}}Med.\\Tok.\end{tabular}} &
    \textbf{\begin{tabular}{@{}c@{}}Med.\\Lat.(s)\end{tabular}} \\
    \midrule
    \multicolumn{6}{c}{\textbf{Qwen3.5-9B}} \\
    \cmidrule{1-6}
    ($\underline{\text{T2}, \text{T1CE}})$ & OFF & 58.9 & 50.5 &  194 & 12.1 \\
    ($\underline{\text{T2}, \text{T1CE}})$ & ON  & 58.9 & 52.5 & 3913 & 173.0 \\
    \addlinespace[4pt]
    $\text{T2} \rightarrow \text{T1CE}$    & OFF & \textbf{75.6} & \textbf{75.5} &  352 & 23.9 \\
    $\text{T2} \rightarrow \text{T1CE}$    & ON  & 60.0 & 55.0 & 5948 & 287.3 \\
    \midrule
    \multicolumn{6}{c}{\textbf{InternVL3.5-8B}} \\
    \cmidrule{1-6}
    ($\underline{\text{T2}, \text{T1CE}})$ & OFF & 54.4 & 48.3 &   72 &  6.1 \\
    ($\underline{\text{T2}, \text{T1CE}})$ & ON  & \textbf{65.6} & \textbf{65.3} & 1926 & 112.0 \\
    \addlinespace[4pt]
    $\text{T2} \rightarrow \text{T1CE}$    & OFF & \textbf{65.6} & 65.2 &  178 & 11.8 \\
    $\text{T2} \rightarrow \text{T1CE}$    & ON  & 58.9 & 58.8 & 3750 & 178.9 \\
    \bottomrule
  \end{tabularx}
\end{table}

\begin{figure}[t]
    \centering
    \includegraphics[width=1\columnwidth]{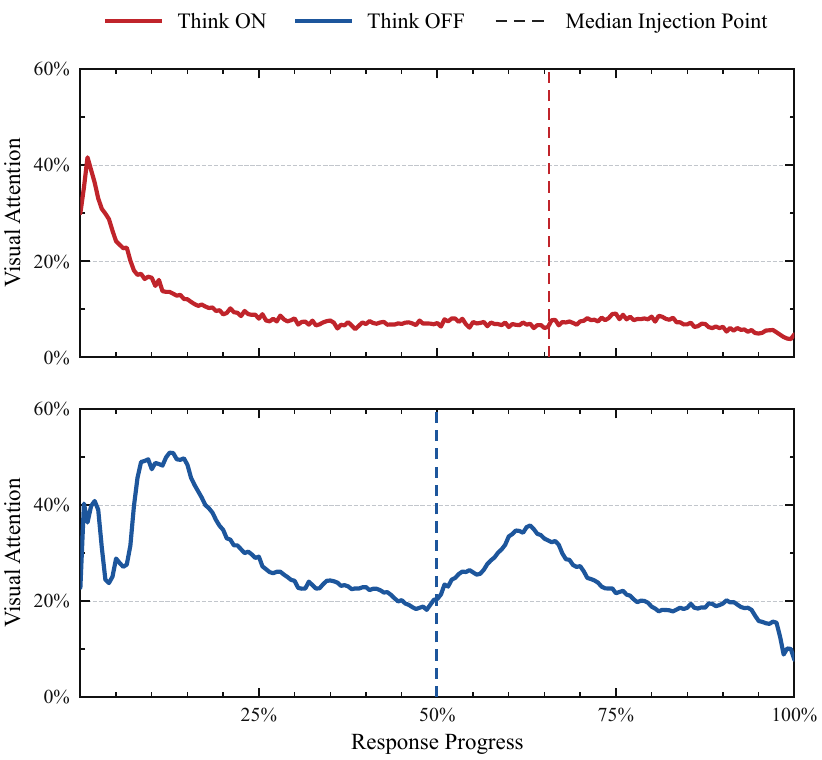}
    \caption{Mean visual attention share across normalized response progress for Qwen3.5-9B along the staged $\text{T2} \rightarrow \text{T1CE}$ pathway (\emph{Think ON} vs.\ \emph{Think OFF}) over paired cases. Vertical dashed lines mark the median point of $\text{T1CE}$ injection.}
    \label{fig:visual_attention}
\end{figure}

\begin{table*}[t]
  \caption{External evaluation on PI-CAI (prostate MRI) and OpenSwissHCC (liver MRI). Best Balanced Accuracy and Macro-F1 per dataset and model are in \textbf{bold}.}
  \label{tab:external_eval}
  \centering
  \small
  \setlength{\tabcolsep}{4.5pt}
  \renewcommand{\arraystretch}{1.15}
  \begin{tabularx}{\textwidth}{ll l l *{4}{>{\centering\arraybackslash}X}}
    \toprule
    & & & & \multicolumn{2}{c}{\textbf{Overall (\%)}} & \multicolumn{2}{c}{\textbf{Per-Class Recall (\%)}} \\
    \cmidrule(lr){5-6} \cmidrule(lr){7-8}
    \textbf{Dataset} & \textbf{Model} & \textbf{Condition} & \textbf{Input Flow} &
    \textbf{\begin{tabular}{@{}c@{}}Balanced\\Accuracy\end{tabular}} &
    \textbf{Macro-F1} &
    \textbf{Positive} & \textbf{Negative} \\
    \midrule
    \multirow{4}{*}{PI-CAI}
      & \multirow{2}{*}{Qwen3.5-9B}
        & Joint & ($\underline{\text{T2W}, \text{ADC}, \text{HBV}})$ & 47.1 & 46.9 & 21.8 & 72.3 \\
      & & \cellcolor{blue!6}\textbf{Staged (Ours)} & \cellcolor{blue!6}$\text{T2W} \rightarrow \text{ADC, HBV}$ & \cellcolor{blue!6}\textbf{57.2} & \cellcolor{blue!6}\textbf{57.4} & \cellcolor{blue!6}27.6 & \cellcolor{blue!6}86.9 \\
      \cmidrule{2-8}
      & \multirow{2}{*}{InternVL3.5-8B}
        & Joint & ($\underline{\text{T2W}, \text{ADC}, \text{HBV}})$ & \textbf{47.2} & 43.4 & 5.7 & 88.7 \\
      & & \cellcolor{blue!6}\textbf{Staged (Ours)} & \cellcolor{blue!6}$\text{T2W} \rightarrow \text{ADC, HBV}$ & \cellcolor{blue!6}46.6 & \cellcolor{blue!6}\textbf{45.4} & \cellcolor{blue!6}13.8 & \cellcolor{blue!6}79.3 \\
    \midrule
    \multirow{4}{*}{OpenSwissHCC}
      & \multirow{2}{*}{Qwen3.5-9B}
        & Joint & ($\underline{\text{Art.}, \text{PV}})$ & 54.8 & 45.1 & 11.1 & 98.6 \\
      & & \cellcolor{blue!6}\textbf{Staged (Ours)} & \cellcolor{blue!6}$\text{Art.} \rightarrow \text{PV}$ & \cellcolor{blue!6}\textbf{61.9} & \cellcolor{blue!6}\textbf{60.7} & \cellcolor{blue!6}41.3 & \cellcolor{blue!6}82.6 \\
      \cmidrule{2-8}
      & \multirow{2}{*}{InternVL3.5-8B}
        & Joint & ($\underline{\text{Art.}, \text{PV}})$ & 50.0 & 34.3 & 0.0 & 100.0 \\
      & & \cellcolor{blue!6}\textbf{Staged (Ours)} & \cellcolor{blue!6}$\text{Art.} \rightarrow \text{PV}$ & \cellcolor{blue!6}\textbf{52.2} & \cellcolor{blue!6}\textbf{47.0} & \cellcolor{blue!6}19.4 & \cellcolor{blue!6}85.1 \\
    \bottomrule
  \end{tabularx}
\end{table*}
\subsection{Primary Comparison on Brain MRI}
\label{sec:res_primary}

\Cref{tab:primary_comparison} highlights a key bottleneck in how frozen VLMs integrate multi-sequence MRI: diagnostic failure under joint presentation appears driven largely by modality dominance rather than missing information. Even in isolation, $\text{T1CE}$ induces a strong metastasis prior across both models---yielding near-ceiling MET recall while severely under-detecting GBM---likely because contrast-enhancing lesions and central necrosis visually overlap across both tumor types. By contrast, $\text{T2}$ highlights peritumoral edema and parenchymal structure without contrast dominance, producing substantially more balanced predictions. Crucially, bundling both sequences together fails to yield synergy; instead, the $\text{T1CE}$ prior tends to override $\text{T2}$ cues, particularly when $\text{T1CE}$ appears second in $(\underline{\text{T2}, \text{T1CE}})$, consistent with a recency bias that pulls overall accuracy even below $\text{T2}$ alone.

Staged inference mitigates this joint interference by separating initial hypothesis formation from secondary integration, consistently improving GBM sensitivity. Yet the two architectures exhibit distinct revision dynamics. Qwen3.5-9B shows strong initial-impression anchoring: starting on $\text{T1CE}$ ($\text{T1CE} \rightarrow \text{T2}$) leaves performance near its $\text{T1CE}$-only baseline, whereas starting on the more balanced $\text{T2}$ sequence ($\text{T2} \rightarrow \text{T1CE}$) allows subsequent $\text{T1CE}$ evidence to refine rather than override the interpretation---lifting balanced accuracy to 75.6\% ($+$16.7 percentage points over its order-matched joint baseline). InternVL3.5-8B instead improves symmetrically to 65.6\% balanced accuracy in both directions, suggesting that temporal separation alone helps relieve cross-sequence interference in this model.

These gains incur roughly double the median latency due to two-stage generation and mid-inference visual prefill. More importantly, they raise a critical mechanistic question: whether the improvement genuinely arises from injecting complementary visual evidence mid-inference, or merely from giving the model extra generation steps to reconsider its output---a distinction we isolate next.

\subsection{Think More or Look Again}
\label{sec:res_text_vs_staged}

To test whether staged gains stem from new visual evidence rather than additional reasoning steps, \Cref{tab:text_vs_staged} contrasts staged injection against text-only continuation (\texttt{Wait} without new images). Across every single-sequence and joint baseline on both models, prolonging generation without fresh visual input consistently degrades performance, collapsing balanced accuracy into the 40.0--53.3\% range---near or below random chance---while consuming substantial inference time (even exceeding staged latency on Qwen3.5-9B). Without new visual evidence to ground revision, forced continuation appears to induce ungrounded self-contradiction, causing models to talk themselves out of valid initial impressions. These results suggest that extra generation steps or explicit hesitation prompting alone do not explain the staged improvement.

Comparing staged inference with and without the \texttt{Wait} cue further points to visual injection as the primary driver, while clarifying its boundary conditions. Along the $\text{T2} \rightarrow \text{T1CE}$ trajectory in \Cref{tab:text_vs_staged}, omitting the cue achieves 76.7\% on Qwen3.5-9B and 67.8\% on InternVL3.5-8B---performing comparably to or slightly above the \texttt{Wait}-prompted runs (a difference of one to two cases at $N=90$). Once a balanced $\text{T2}$ representation is formed, introducing $\text{T1CE}$ directly appears sufficient without verbal nudging. However, this does not rescue an unfavorable initial anchor on Qwen3.5-9B: under $\text{T1CE} \rightarrow \text{T2}$, cue-free injection dips to 60.0\% (below the 64.4\% $\text{T1CE}$ baseline) and adding \texttt{Wait} merely restores baseline performance, whereas InternVL3.5-8B preserves its gains over joint baselines regardless of cueing. An extended ablation with an attention-directing cue (\texttt{New evidence}) in \Cref{sec:supp_cue_ablation} shows a similar trend, yielding no consistent advantage over cue-free injection (reducing Qwen3.5-9B on $\text{T2} \rightarrow \text{T1CE}$ while modestly improving InternVL3.5-8B), while \Cref{sec:supp_class_balance} summarizes the accompanying class-prediction shifts under text-only continuation.

\subsection{Case-Level Revision Dynamics}
\label{sec:res_paired}

Aggregate accuracy cannot separate genuine error correction from prediction shuffling. \Cref{tab:paired_vs_joint} therefore tracks paired transitions against order-matched joint baselines, separating constructive fixes ($\text{W}\rightarrow\text{C}$) from regressions ($\text{C}\rightarrow\text{W}$). This breakdown illuminates the behavior behind the text-only drop: prompting \texttt{Wait} without new images does not leave the model anchored on its prior guess, but is accompanied by widespread label churn---flipping roughly half to three-quarters of patients, with regressions consistently outnumbering fixes (net $-5$ to $-19$ cases) and overturning valid diagnoses faster than correcting mistakes.

Staged visual injection shifts this behavior from erratic churn to selective, net-positive updating. Turnover drops markedly, and along $\text{T2} \rightarrow \text{T1CE}$, fixes outnumber regressions by more than two-to-one ($+$15 on Qwen3.5-9B; $+$10 on InternVL3.5-8B). Crucially, revision is not monotonic: 9--10 initially correct joint predictions still regress, indicating that overall gains reflect a favorable correction--regression balance rather than error-free updating. Directional asymmetry is likewise visible: under $\text{T1CE} \rightarrow \text{T2}$, InternVL3.5-8B keeps a modest net gain ($+$5) while Qwen3.5-9B trades $12$-for-$12$ (zero net change), suggesting that an unfavorable starting sequence reshuffles predictions without net diagnostic benefit. Full paired transition counts, significance tests, and 95\% confidence intervals are provided in \Cref{sec:supp_stats}.

\subsection{Built-in Thinking vs.\ Staged Visual Inference}
\label{sec:res_thinking}

\Cref{tab:thinking_ablation} evaluates whether built-in thinking can substitute for, or enhance, $\text{T2}$-first staging. First, \emph{Think OFF} staging matches or outperforms \emph{Think ON} joint inference at a fraction of the compute: on Qwen3.5-9B, thinking on $(\underline{\text{T2}, \text{T1CE}})$ inflates tokens roughly twentyfold with zero balanced-accuracy gain (58.9\% vs.\ 75.6\% staged), while on InternVL3.5-8B thinking helps resolve joint inputs (lifting $(\underline{\text{T2}, \text{T1CE}})$ to 65.6\%) yet \emph{Think OFF} staging matches that accuracy with roughly $10\times$ fewer tokens and latency. Second, pairing $\text{T2} \rightarrow \text{T1CE}$ staging with \emph{Think ON} degrades both models in \Cref{tab:thinking_ablation} (to 60.0\% and 58.9\%) while expanding outputs to thousands of tokens. Broader comparisons across both sequence orders and cue settings in \Cref{sec:supp_thinking_orders} show that \emph{Think ON} consistently underperforms concise staging on Qwen3.5-9B and exhibits sensitivity to pre-injection chain length and cueing on InternVL3.5-8B at roughly tenfold higher compute cost.

\Cref{fig:visual_attention} compares visual-token attention trajectories across both conditions on Qwen3.5-9B. Under \emph{Think ON}, visual attention falls early in the extended chain into a low plateau ($<$10\%) with little rebound when $\text{T1CE}$ arrives roughly two-thirds into generation. Under \emph{Think OFF}, by contrast, the model maintains a higher visual attention share and shows a clear upward shift immediately after midpoint $\text{T1CE}$ injection. These observational patterns suggest that prolonged internal reasoning shifts focus away from visual inputs as self-generated text accumulates, whereas concise decoding preserves stronger engagement with newly introduced images; corresponding attention budgets and InternVL3.5-8B trajectories are detailed in \Cref{sec:supp_attention_budget}.

\subsection{External Transfer to Prostate and Liver MRI}
\label{sec:res_external}

Most existing general-purpose VLM benchmarks focus on single-image perception or synthetic multi-image visual question answering (VQA) rather than real-world tasks requiring genuinely complementary acquisitions of the same physical subject. To evaluate whether staged presentation transfers beyond neuro-oncology in such settings, \Cref{tab:external_eval} reports zero-shot generalization on two public multi-sequence MRI cohorts using full, unmasked volumes without clinical metadata. On PI-CAI, the task is case-level detection of csPCa (Gleason grade group $\ge 2$) versus benign or indolent findings; staging $\text{T2W} \rightarrow \text{ADC, HBV}$ presents zonal prostate anatomy first before introducing diffusion-restriction cues. On OpenSwissHCC, the task is identifying HCC in patients with chronic liver disease---a challenging setting where background cirrhosis and benign nodules frequently confound diagnosis---and staging $\text{Art.} \rightarrow \text{PV}$ aligns with clinical contrast assessment by presenting arterial enhancement prior to portal-venous washout.

Across both external tasks, joint presentation again exhibits severe single-class collapse, here skewing heavily toward negative predictions (with positive sensitivity falling to 0--22\% under joint inputs). Staged inference mitigates this bias most consistently on Qwen3.5-9B. On PI-CAI, staging raises balanced accuracy from 47.1\% to 57.2\% and Macro-F1 to 57.4\% ($+$10.1 and $+$10.5 points), improving both positive and negative recall. On OpenSwissHCC, staging improves balanced accuracy from 54.8\% to 61.9\% and Macro-F1 by $+$15.6 points (to 60.7\%) by nearly quadrupling positive sensitivity. On InternVL3.5-8B, staging breaks total negative-class collapse on OpenSwissHCC (lifting Macro-F1 by $+$12.7 points as positive recall rises from 0.0\% to 19.4\%, with balanced accuracy at 52.2\% over the 129/132 valid examinations; \Cref{sec:supp_stats}), whereas PI-CAI remains near chance (46.6\% vs.\ 47.2\% balanced accuracy).

Together, these findings delineate both the potential and the practical boundary conditions of inference-time staging. When a frozen VLM retains usable sensitivity to the underlying imaging cues---as Qwen3.5-9B exhibits on both prostate and liver MRI---decoupling sequence presentation consistently mitigates joint-input collapse and yields meaningful relative gains. Conversely, InternVL3.5-8B's near-chance performance on PI-CAI highlights a clear boundary: staging can reorganize and unlock visual evidence a frozen backbone already perceives, but cannot compensate when subtle focal lesions across unlocalized 3D studies fall below baseline perceptual sensitivity.

\subsection{Mechanistic and Exploratory Audit}
\label{sec:res_mechanistic_audit}

To examine the internal behavior underlying these results, \Cref{sec:supp_mechanistic} presents a mechanistic and exploratory audit with full visualizations on the brain MRI cohort. Two consistent observations emerge. First, under \emph{Think OFF}, both models actively attend to whichever sequence is injected second, indicating that Qwen3.5-9B's weaker performance under $\text{T1CE} \rightarrow \text{T2}$ does not stem from simply neglecting late $\text{T2}$ frames---whereas enabling \emph{Think ON} shifts the majority of context attention onto self-generated text. At the case level, $\text{T2} \rightarrow \text{T1CE}$ staging primarily recovers GBM cases masked under joint $\text{T1CE}$ dominance, in contrast to text-only continuation which predominantly overturns valid MET calls.

Second, diagnostic errors under joint inputs do not simply reflect a failure to attend to tumor-bearing slices. Qwen3.5-9B already assigns higher raw attention to tumor slices across both correct and misclassified cases, whereas InternVL3.5-8B still benefits from staging despite pronounced opening-slice positional sinks and weak tumor-slice separation. Together, these observations suggest that the multi-sequence bottleneck lies less in whether the model notices relevant frames or slices, and more in how complementary sequences are ordered and integrated during inference.

\section{Conclusion}
\label{sec:conclusion}

We studied staged visual inference, a training-free test-time strategy that organizes when complementary imaging sequences enter frozen VLMs. On multi-sequence brain MRI, presenting T2 before T1CE substantially improves diagnostic performance over joint inputs, whereas text-only continuation destabilizes predictions and built-in thinking adds heavy compute overhead without reliable gains. Evaluations on prostate and liver MRI further show that staging can mitigate single-class collapse across clinical tasks without retraining or architectural modifications. Overall, decoupling how complementary visual evidence enters inference offers a simple and effective paradigm for test-time scaling in multimodal medical reasoning.

{
    \small
    \setlength{\bibsep}{-0.6pt}
    \bibliographystyle{ieeenat_fullname}
    \bibliography{main}
}

\clearpage
\appendix
\crefalias{section}{appendix}
\crefalias{subsection}{appendix}

\section{Extended Dataset Details and Preprocessing}
\label{sec:supp_datasets}

To evaluate training-free staged visual inference across diverse clinical domains, acquisition protocols, and contamination regimes, we consider three multi-sequence volumetric MRI cohorts spanning neuro-oncology, prostate imaging, and liver imaging. 

All three datasets are processed and presented under a unified, label-blind evaluation protocol: frozen models receive complete axial slice sequences directly in anatomical order without subsampling, spatial lesion masks, bounding boxes, or clinical metadata.

\subsection{Cohort Specifications}
\label{sec:supp_cohorts}

\paragraph{Brain MRI (Primary Cohort).}
Our primary cohort comprises 90 retrospectively collected patients balanced equally between GBM ($n=45$) and MET ($n=45$). Examinations were acquired during routine clinical workflows across four academic hospitals using Siemens MRI scanners, totaling 4,091 axial slices across paired T1CE and T2 sequences. 

All diagnostic labels were established strictly via histopathological confirmation from surgical resection or biopsy specimens rather than preoperative radiological impressions. Because this clinical dataset was independently curated from institutional records rather than public repositories, it minimizes the risk of pretraining data contamination.

\paragraph{PI-CAI (Prostate MRI).}
To evaluate multi-sequence generalization in a three-sequence diagnostic setting, we use the public PI-CAI training and development archive~\cite{saha2023artificial}, acquired across three Dutch tertiary centers on 1.5T and 3T scanners (Siemens Healthineers and Philips Medical Systems). To ensure an unbiased, predefined test split without ad-hoc case selection, we evaluate the entire official \texttt{fold~0} partition ($N=300$ examinations: 87 csPCa [Gleason grade group $\ge 2$] and 213 benign or indolent cases [Gleason grade group $\le 1$], confirmed by histopathology and/or $\ge 3$-year clinical follow-up). Every examination includes paired axial T2W, ADC, and HBV sequences.

\paragraph{OpenSwissHCC (Liver MRI).}
To evaluate generalization on a newly released, low-contamination cohort, we test on OpenSwissHCC \cite{tomography12090133}, a multiphasic liver MRI resource released in 2026 comprising 132 adult patients with chronic liver disease scanned on 1.5T and 3T systems (63 HCC-positive and 69 HCC-negative examinations). Released contemporaneously with or after the evaluated VLM checkpoints, OpenSwissHCC reduces pretraining memorization risk while presenting a challenging diagnostic background of cirrhosis and benign regenerative nodules. For every examination, we pair the water-only T1-weighted Art.\ and PV volumes (using the clinically designated arterial phase recorded in the cohort metadata when multiple dynamic timing-test captures are present).

\subsection{Rationale for Sequence Complementarity}
\label{sec:supp_clinical_defs}

Across all three tasks, reliable diagnosis requires integrating complementary imaging characteristics distributed across distinct MRI acquisitions rather than relying on any single sequence alone:
\vspace{0.5cm} 
\begin{itemize}
\setlength\itemsep{0.5cm}
    \item \textbf{Brain MRI (T2 and T1CE):} Differentiating GBM from brain MET on structural MRI is clinically challenging because both frequently show ring enhancement and central necrosis. T2 highlights tissue water content, parenchymal mass effect, and peritumoral edema extending beyond the tumor margin, whereas T1CE captures gadolinium leakage across a disrupted blood--brain barrier, delineating enhancing tumor walls and necrotic cores.
    \item \textbf{Prostate MRI (T2W, ADC, and HBV):} T2W provides high-resolution zonal anatomy for localizing the peripheral and transition zones, while paired diffusion acquisitions (ADC maps and HBV images) capture restricted water diffusion characteristic of densely packed, clinically significant malignancies.
    \item \textbf{Liver MRI (Art.\ and PV):} Multiphasic HCC assessment relies on hemodynamic contrast dynamics across phases: Art.\ imaging captures arterial phase hyperenhancement of hypervascular lesions, whereas PV imaging reveals subsequent contrast washout relative to surrounding cirrhotic parenchyma.
\end{itemize}

\begin{table*}[t]
  \caption{Cohort composition and per-sequence axial slice counts across the primary and external multi-sequence MRI evaluations.}
  \centering
  \small
  \setlength{\tabcolsep}{5pt}
  \renewcommand{\arraystretch}{1.12}
  \begin{tabularx}{\textwidth}{l l c c l >{\centering\arraybackslash}X}
    \toprule
    \textbf{Dataset} & \textbf{Clinical Task} & \textbf{Cases (Pos / Neg)} & \textbf{Field Strength} &
    \textbf{Sequence} & \textbf{Slices / Volume (Min--Max)} \\
    \midrule
    \multirow{2}{*}{Brain MRI (Primary)}
      & \multirow{2}{*}{GBM vs.\ MET}
      & \multirow{2}{*}{90 (45 / 45)}
      & \multirow{2}{*}{1.5T}
      & T1CE & 15--160 \\
      & & & & T2   & 15--62  \\
    \midrule
    \multirow{3}{*}{PI-CAI (Fold 0)}
      & \multirow{3}{*}{csPCa vs.\ Negative}
      & \multirow{3}{*}{300 (87 / 213)}
      & \multirow{3}{*}{1.5T / 3T}
      & T2W & 17--33 \\
      & & & & ADC & 16--33 \\
      & & & & HBV & 16--33 \\
    \midrule
    \multirow{2}{*}{OpenSwissHCC}
      & \multirow{2}{*}{HCC+ vs.\ HCC$-$}
      & \multirow{2}{*}{132 (63 / 69)}
      & \multirow{2}{*}{1.5T / 3T}
      & Art. & 48--104 \\
      & & & & PV   & 48--104 \\
    \bottomrule
  \end{tabularx}

  \label{tab:supp_dataset_summary}
\end{table*}

\subsection{Image Conversion and Presentation Protocol}
\label{sec:supp_preprocessing}

All three cohorts were prepared and supplied to the frozen VLMs through an identical volumetric export and input-formatting pipeline, ensuring that cross-dataset comparisons reflect diagnostic reasoning behavior rather than differences in preprocessing:

\paragraph{Volumetric-to-2D Export.}
Original 3D volumes were reoriented to canonical RAS anatomical coordinates and stripped of all patient identifiers and metadata headers before axial slicing. To preserve soft-tissue contrast and avoid dynamic-range compression from bright vascular or coil artifacts, voxel intensities were clipped per volume between the 0.5th and 99.5th percentiles of finite values, linearly rescaled to 8-bit integers ($[0, 255]$), and exported sequentially along the axial plane as high-quality JPEG images (quality $= 98$).

\paragraph{Full-Volume Sequencing and Independent Framing.}
Across all cohorts and conditions, every exported slice from each selected sequence was loaded in RGB format and passed in ascending anatomical order without slice filtering, cropping, or subsampling (\Cref{tab:supp_dataset_summary}). Each slice was presented as an independent visual frame rather than merged into multi-slice grid mosaics, preserving native per-slice resolution and spatial structure. Immediately before each sequence's frames, a concise sequence-name header was inserted so the model could distinguish acquisition phases without access to patient identifiers or clinical annotations.

\begin{table*}[t]
  \caption{Cue-framing ablation for staged visual inference (\emph{Think OFF}) on the brain MRI cohort (\texttt{New evid.} denotes the \texttt{New evidence} cue). Best Balanced Accuracy and Macro-F1 per model are in \textbf{bold}.}
  \centering
  \small
  \setlength{\tabcolsep}{4pt}
  \renewcommand{\arraystretch}{1.12}
  \begin{tabularx}{\textwidth}{l l c *{8}{>{\centering\arraybackslash}X}}
    \toprule
    & & & \multicolumn{2}{c}{\textbf{Overall (\%)}} & \multicolumn{2}{c}{\textbf{Recall (\%)}} & \multicolumn{3}{c}{\textbf{Med.\ Tokens}} & \\
    \cmidrule(lr){4-5} \cmidrule(lr){6-7} \cmidrule(lr){8-10}
    \textbf{Model} & \textbf{Input Flow} & \textbf{Cue} &
    \textbf{\begin{tabular}{@{}c@{}}Bal.\\Acc.\end{tabular}} &
    \textbf{\begin{tabular}{@{}c@{}}Macro-\\F1\end{tabular}} &
    \textbf{GBM} & \textbf{MET} &
    \textbf{Pre} & \textbf{Post} & \textbf{Tot.} &
    \textbf{\begin{tabular}{@{}c@{}}Lat.\\(s)\end{tabular}} \\
    \midrule
    \multirow{6}{*}{Qwen3.5-9B}
      & \cellcolor{blue!6}$\text{T1CE} \rightarrow \text{T2}$ & \cellcolor{blue!6}\textemdash           & \cellcolor{blue!6}60.0 & \cellcolor{blue!6}55.7 & \cellcolor{blue!6}28.9 & \cellcolor{blue!6}91.1 & \cellcolor{blue!6}172 & \cellcolor{blue!6}170 & \cellcolor{blue!6}346 & \cellcolor{blue!6}20.6 \\
      & \cellcolor{blue!6}$\text{T1CE} \rightarrow \text{T2}$ & \cellcolor{blue!6}\texttt{Wait}      & \cellcolor{blue!6}64.4 & \cellcolor{blue!6}63.0 & \cellcolor{blue!6}44.4 & \cellcolor{blue!6}84.4 & \cellcolor{blue!6}172 & \cellcolor{blue!6}182 & \cellcolor{blue!6}356 & \cellcolor{blue!6}21.8 \\
      & \cellcolor{blue!6}$\text{T1CE} \rightarrow \text{T2}$ & \cellcolor{blue!6}\texttt{New evid.} & \cellcolor{blue!6}60.0 & \cellcolor{blue!6}56.9 & \cellcolor{blue!6}33.3 & \cellcolor{blue!6}86.7 & \cellcolor{blue!6}172 & \cellcolor{blue!6}206 & \cellcolor{blue!6}378 & \cellcolor{blue!6}24.2 \\
      & \cellcolor{blue!6}$\text{T2} \rightarrow \text{T1CE}$ & \cellcolor{blue!6}\textemdash           & \cellcolor{blue!6}\textbf{76.7} & \cellcolor{blue!6}\textbf{76.7} & \cellcolor{blue!6}75.6 & \cellcolor{blue!6}77.8 & \cellcolor{blue!6}174 & \cellcolor{blue!6}172 & \cellcolor{blue!6}350 & \cellcolor{blue!6}21.8 \\
      & \cellcolor{blue!6}$\text{T2} \rightarrow \text{T1CE}$ & \cellcolor{blue!6}\texttt{Wait}      & \cellcolor{blue!6}75.6 & \cellcolor{blue!6}75.5 & \cellcolor{blue!6}71.1 & \cellcolor{blue!6}80.0 & \cellcolor{blue!6}174 & \cellcolor{blue!6}176 & \cellcolor{blue!6}352 & \cellcolor{blue!6}23.9 \\
      & \cellcolor{blue!6}$\text{T2} \rightarrow \text{T1CE}$ & \cellcolor{blue!6}\texttt{New evid.} & \cellcolor{blue!6}71.1 & \cellcolor{blue!6}71.1 & \cellcolor{blue!6}66.7 & \cellcolor{blue!6}75.6 & \cellcolor{blue!6}174 & \cellcolor{blue!6}194 & \cellcolor{blue!6}369 & \cellcolor{blue!6}24.4 \\
    \midrule
    \multirow{6}{*}{InternVL3.5-8B}
      & \cellcolor{blue!6}$\text{T1CE} \rightarrow \text{T2}$ & \cellcolor{blue!6}\textemdash           & \cellcolor{blue!6}64.4 & \cellcolor{blue!6}64.4 & \cellcolor{blue!6}60.0 & \cellcolor{blue!6}68.9 & \cellcolor{blue!6}80 & \cellcolor{blue!6}93 & \cellcolor{blue!6}173 & \cellcolor{blue!6}11.7 \\
      & \cellcolor{blue!6}$\text{T1CE} \rightarrow \text{T2}$ & \cellcolor{blue!6}\texttt{Wait}      & \cellcolor{blue!6}65.6 & \cellcolor{blue!6}65.5 & \cellcolor{blue!6}62.2 & \cellcolor{blue!6}68.9 & \cellcolor{blue!6}80 & \cellcolor{blue!6}92 & \cellcolor{blue!6}169 & \cellcolor{blue!6}11.9 \\
      & \cellcolor{blue!6}$\text{T1CE} \rightarrow \text{T2}$ & \cellcolor{blue!6}\texttt{New evid.} & \cellcolor{blue!6}66.7 & \cellcolor{blue!6}66.7 & \cellcolor{blue!6}66.7 & \cellcolor{blue!6}66.7 & \cellcolor{blue!6}80 & \cellcolor{blue!6}93 & \cellcolor{blue!6}174 & \cellcolor{blue!6}11.8 \\
      & \cellcolor{blue!6}$\text{T2} \rightarrow \text{T1CE}$ & \cellcolor{blue!6}\textemdash           & \cellcolor{blue!6}67.8 & \cellcolor{blue!6}67.3 & \cellcolor{blue!6}55.6 & \cellcolor{blue!6}80.0 & \cellcolor{blue!6}84 & \cellcolor{blue!6}95 & \cellcolor{blue!6}180 & \cellcolor{blue!6}12.1 \\
      & \cellcolor{blue!6}$\text{T2} \rightarrow \text{T1CE}$ & \cellcolor{blue!6}\texttt{Wait}      & \cellcolor{blue!6}65.6 & \cellcolor{blue!6}65.2 & \cellcolor{blue!6}55.6 & \cellcolor{blue!6}75.6 & \cellcolor{blue!6}84 & \cellcolor{blue!6}94 & \cellcolor{blue!6}178 & \cellcolor{blue!6}11.8 \\
      & \cellcolor{blue!6}$\text{T2} \rightarrow \text{T1CE}$ & \cellcolor{blue!6}\texttt{New evid.} & \cellcolor{blue!6}\textbf{68.9} & \cellcolor{blue!6}\textbf{68.5} & \cellcolor{blue!6}57.8 & \cellcolor{blue!6}80.0 & \cellcolor{blue!6}84 & \cellcolor{blue!6}95 & \cellcolor{blue!6}178 & \cellcolor{blue!6}11.9 \\
    \bottomrule
  \end{tabularx}

  \label{tab:supp_cue_ablation}
\end{table*}

\begin{table*}[t]
  \caption{Paired \emph{Think OFF} versus \emph{Think ON} comparison across both sequence orders and cue settings on the brain MRI cohort. Best Balanced Accuracy and Macro-F1 per model are in \textbf{bold}.}
  \centering
  \small
  \setlength{\tabcolsep}{3.2pt}
  \renewcommand{\arraystretch}{1.12}
  \begin{tabularx}{\textwidth}{l l c c *{8}{>{\centering\arraybackslash}X}}
    \toprule
    & & & & \multicolumn{2}{c}{\textbf{Overall (\%)}} & \multicolumn{2}{c}{\textbf{Recall (\%)}} & \multicolumn{3}{c}{\textbf{Med.\ Tokens}} & \\
    \cmidrule(lr){5-6} \cmidrule(lr){7-8} \cmidrule(lr){9-11}
    \textbf{Cond.} & \textbf{Input Flow} & \textbf{Cue} & \textbf{Think} &
    \textbf{\begin{tabular}{@{}c@{}}Bal.\\Acc.\end{tabular}} &
    \textbf{\begin{tabular}{@{}c@{}}Macro-\\F1\end{tabular}} &
    \textbf{GBM} & \textbf{MET} &
    \textbf{Pre} & \textbf{Post} & \textbf{Tot.} &
    \textbf{\begin{tabular}{@{}c@{}}Lat.\\(s)\end{tabular}} \\
    \midrule
    \multicolumn{12}{c}{\textbf{Qwen3.5-9B}} \\
    \midrule
    \multirow{4}{*}{Joint}
      & $(\underline{\text{T1CE}, \text{T2}})$ & \textemdash & OFF & 64.4 & 60.0 & 31.1 & 97.8  & \textemdash & \textemdash & 182  & 12.7 \\
      & $(\underline{\text{T1CE}, \text{T2}})$ & \textemdash & ON  & 62.2 & 58.8 & 33.3 & 91.1  & \textemdash & \textemdash & 3118 & 137.7 \\
      & $(\underline{\text{T2}, \text{T1CE}})$ & \textemdash & OFF & 58.9 & 50.5 & 17.8 & 100.0 & \textemdash & \textemdash & 194  & 12.1 \\
      & $(\underline{\text{T2}, \text{T1CE}})$ & \textemdash & ON  & 58.9 & 52.5 & 22.2 & 95.6  & \textemdash & \textemdash & 3913 & 173.0 \\
    \midrule
    \cellcolor{blue!6} & \cellcolor{blue!6}$\text{T1CE} \rightarrow \text{T2}$ & \cellcolor{blue!6}\textemdash    & \cellcolor{blue!6}OFF & \cellcolor{blue!6}60.0 & \cellcolor{blue!6}55.7 & \cellcolor{blue!6}28.9 & \cellcolor{blue!6}91.1 & \cellcolor{blue!6}172  & \cellcolor{blue!6}170  & \cellcolor{blue!6}346  & \cellcolor{blue!6}20.6 \\
    \cellcolor{blue!6} & \cellcolor{blue!6}$\text{T1CE} \rightarrow \text{T2}$ & \cellcolor{blue!6}\textemdash    & \cellcolor{blue!6}ON  & \cellcolor{blue!6}61.1 & \cellcolor{blue!6}55.9 & \cellcolor{blue!6}26.7 & \cellcolor{blue!6}95.6 & \cellcolor{blue!6}2448 & \cellcolor{blue!6}728  & \cellcolor{blue!6}3936 & \cellcolor{blue!6}174.8 \\
    \cellcolor{blue!6} & \cellcolor{blue!6}$\text{T1CE} \rightarrow \text{T2}$ & \cellcolor{blue!6}\texttt{Wait} & \cellcolor{blue!6}OFF & \cellcolor{blue!6}64.4 & \cellcolor{blue!6}63.0 & \cellcolor{blue!6}44.4 & \cellcolor{blue!6}84.4 & \cellcolor{blue!6}172  & \cellcolor{blue!6}182  & \cellcolor{blue!6}356  & \cellcolor{blue!6}21.8 \\
    \cellcolor{blue!6} & \cellcolor{blue!6}$\text{T1CE} \rightarrow \text{T2}$ & \cellcolor{blue!6}\texttt{Wait} & \cellcolor{blue!6}ON  & \cellcolor{blue!6}64.4 & \cellcolor{blue!6}61.2 & \cellcolor{blue!6}35.6 & \cellcolor{blue!6}93.3 & \cellcolor{blue!6}2634 & \cellcolor{blue!6}1615 & \cellcolor{blue!6}4762 & \cellcolor{blue!6}208.9 \\
    \cellcolor{blue!6} & \cellcolor{blue!6}$\text{T2} \rightarrow \text{T1CE}$ & \cellcolor{blue!6}\textemdash    & \cellcolor{blue!6}OFF & \cellcolor{blue!6}\textbf{76.7} & \cellcolor{blue!6}\textbf{76.7} & \cellcolor{blue!6}75.6 & \cellcolor{blue!6}77.8 & \cellcolor{blue!6}174  & \cellcolor{blue!6}172  & \cellcolor{blue!6}350  & \cellcolor{blue!6}21.8 \\
    \cellcolor{blue!6} & \cellcolor{blue!6}$\text{T2} \rightarrow \text{T1CE}$ & \cellcolor{blue!6}\textemdash    & \cellcolor{blue!6}ON  & \cellcolor{blue!6}57.8 & \cellcolor{blue!6}50.7 & \cellcolor{blue!6}20.0 & \cellcolor{blue!6}95.6 & \cellcolor{blue!6}3564 & \cellcolor{blue!6}1208 & \cellcolor{blue!6}5422 & \cellcolor{blue!6}234.9 \\
    \cellcolor{blue!6} & \cellcolor{blue!6}$\text{T2} \rightarrow \text{T1CE}$ & \cellcolor{blue!6}\texttt{Wait} & \cellcolor{blue!6}OFF & \cellcolor{blue!6}75.6 & \cellcolor{blue!6}75.5 & \cellcolor{blue!6}71.1 & \cellcolor{blue!6}80.0 & \cellcolor{blue!6}174  & \cellcolor{blue!6}176  & \cellcolor{blue!6}352  & \cellcolor{blue!6}23.9 \\
    \cellcolor{blue!6}\multirow{-8}{*}{\textbf{Staged}}
      & \cellcolor{blue!6}$\text{T2} \rightarrow \text{T1CE}$ & \cellcolor{blue!6}\texttt{Wait} & \cellcolor{blue!6}ON  & \cellcolor{blue!6}60.0 & \cellcolor{blue!6}55.0 & \cellcolor{blue!6}26.7 & \cellcolor{blue!6}93.3 & \cellcolor{blue!6}3666 & \cellcolor{blue!6}2131 & \cellcolor{blue!6}5948 & \cellcolor{blue!6}287.3 \\
    \midrule
    \multicolumn{12}{c}{\textbf{InternVL3.5-8B}} \\
    \midrule
    \multirow{4}{*}{Joint}
      & $(\underline{\text{T1CE}, \text{T2}})$ & \textemdash & OFF & 60.0 & 54.2 & 24.4 & 95.6 & \textemdash & \textemdash & 78   & 6.7 \\
      & $(\underline{\text{T1CE}, \text{T2}})$ & \textemdash & ON  & 64.4 & 64.2 & 55.6 & 73.3 & \textemdash & \textemdash & 1956 & 104.9 \\
      & $(\underline{\text{T2}, \text{T1CE}})$ & \textemdash & OFF & 54.4 & 48.3 & 20.0 & 88.9 & \textemdash & \textemdash & 72   & 6.1 \\
      & $(\underline{\text{T2}, \text{T1CE}})$ & \textemdash & ON  & 65.6 & 65.3 & 57.8 & 73.3 & \textemdash & \textemdash & 1926 & 112.0 \\
    \midrule
    \cellcolor{blue!6} & \cellcolor{blue!6}$\text{T1CE} \rightarrow \text{T2}$ & \cellcolor{blue!6}\textemdash    & \cellcolor{blue!6}OFF & \cellcolor{blue!6}64.4 & \cellcolor{blue!6}64.4 & \cellcolor{blue!6}60.0 & \cellcolor{blue!6}68.9 & \cellcolor{blue!6}80   & \cellcolor{blue!6}93  & \cellcolor{blue!6}173  & \cellcolor{blue!6}11.7 \\
    \cellcolor{blue!6} & \cellcolor{blue!6}$\text{T1CE} \rightarrow \text{T2}$ & \cellcolor{blue!6}\textemdash    & \cellcolor{blue!6}ON  & \cellcolor{blue!6}66.7 & \cellcolor{blue!6}66.4 & \cellcolor{blue!6}57.8 & \cellcolor{blue!6}75.6 & \cellcolor{blue!6}1052 & \cellcolor{blue!6}174 & \cellcolor{blue!6}1581 & \cellcolor{blue!6}72.4 \\
    \cellcolor{blue!6} & \cellcolor{blue!6}$\text{T1CE} \rightarrow \text{T2}$ & \cellcolor{blue!6}\texttt{Wait} & \cellcolor{blue!6}OFF & \cellcolor{blue!6}65.6 & \cellcolor{blue!6}65.5 & \cellcolor{blue!6}62.2 & \cellcolor{blue!6}68.9 & \cellcolor{blue!6}80   & \cellcolor{blue!6}92  & \cellcolor{blue!6}169  & \cellcolor{blue!6}11.9 \\
    \cellcolor{blue!6} & \cellcolor{blue!6}$\text{T1CE} \rightarrow \text{T2}$ & \cellcolor{blue!6}\texttt{Wait} & \cellcolor{blue!6}ON  & \cellcolor{blue!6}\textbf{71.1} & \cellcolor{blue!6}\textbf{70.8} & \cellcolor{blue!6}60.0 & \cellcolor{blue!6}82.2 & \cellcolor{blue!6}1433 & \cellcolor{blue!6}563 & \cellcolor{blue!6}2064 & \cellcolor{blue!6}110.0 \\
    \cellcolor{blue!6} & \cellcolor{blue!6}$\text{T2} \rightarrow \text{T1CE}$ & \cellcolor{blue!6}\textemdash    & \cellcolor{blue!6}OFF & \cellcolor{blue!6}67.8 & \cellcolor{blue!6}67.3 & \cellcolor{blue!6}55.6 & \cellcolor{blue!6}80.0 & \cellcolor{blue!6}84   & \cellcolor{blue!6}95  & \cellcolor{blue!6}180  & \cellcolor{blue!6}12.1 \\
    \cellcolor{blue!6} & \cellcolor{blue!6}$\text{T2} \rightarrow \text{T1CE}$ & \cellcolor{blue!6}\textemdash    & \cellcolor{blue!6}ON  & \cellcolor{blue!6}67.8 & \cellcolor{blue!6}67.8 & \cellcolor{blue!6}66.7 & \cellcolor{blue!6}68.9 & \cellcolor{blue!6}2888 & \cellcolor{blue!6}194 & \cellcolor{blue!6}3400 & \cellcolor{blue!6}149.3 \\
    \cellcolor{blue!6} & \cellcolor{blue!6}$\text{T2} \rightarrow \text{T1CE}$ & \cellcolor{blue!6}\texttt{Wait} & \cellcolor{blue!6}OFF & \cellcolor{blue!6}65.6 & \cellcolor{blue!6}65.2 & \cellcolor{blue!6}55.6 & \cellcolor{blue!6}75.6 & \cellcolor{blue!6}84   & \cellcolor{blue!6}94  & \cellcolor{blue!6}178  & \cellcolor{blue!6}11.8 \\
    \cellcolor{blue!6}\multirow{-8}{*}{\textbf{Staged}}
      & \cellcolor{blue!6}$\text{T2} \rightarrow \text{T1CE}$ & \cellcolor{blue!6}\texttt{Wait} & \cellcolor{blue!6}ON  & \cellcolor{blue!6}58.9 & \cellcolor{blue!6}58.8 & \cellcolor{blue!6}64.4 & \cellcolor{blue!6}53.3 & \cellcolor{blue!6}2646 & \cellcolor{blue!6}826 & \cellcolor{blue!6}3750 & \cellcolor{blue!6}178.9 \\
    \bottomrule
  \end{tabularx}

  \label{tab:supp_thinking_full}
\end{table*}

\section{Prompts and Intervention Protocol}
\label{sec:supp_protocol}

\subsection{Diagnostic Prompts and Continuation Cues}
\label{sec:supp_prompts}

Across all three cohorts, each imaging sequence is preceded by a concise text header identifying the sequence name (e.g., T1-weighted post-contrast MRI, T2-weighted MRI, ADC, high-$b$-value diffusion-weighted imaging (DWI), arterial-phase T1-weighted MRI, or portal-venous-phase T1-weighted MRI). Models receive a standardized task instruction specifying the candidate labels, requesting one brief explanatory paragraph, and enforcing a final answer line. The exact prompts for the brain, prostate, and liver tasks are shown below.

\begin{tcolorbox}[
    fonttitle=\footnotesize\bfseries,
    colback=gray!5!white,
    colframe=black!60,
    left=4pt,
    right=4pt,
    top=3pt,
    bottom=3pt,
    title=Brain MRI Diagnostic Prompt (GBM vs.\ MET),
    fontupper=\fontsize{8pt}{9.5pt}\selectfont,
]
Review the brain MRI images as one volumetric case.

\smallskip
Choose the more likely diagnosis:\\

- Metastasis\\
- Glioblastoma

\smallskip
Give one brief paragraph stating the main distinguishing evidence.

\smallskip

End with exactly one line:\\

Final Answer: Metastasis\\
or\\
Final Answer: Glioblastoma
\end{tcolorbox}

\begin{tcolorbox}[
    fonttitle=\footnotesize\bfseries,
    colback=gray!5!white,
    colframe=black!60,
    left=4pt,
    right=4pt,
    top=3pt,
    bottom=3pt,
    title=Prostate MRI Diagnostic Prompt (csPCa vs.\ Non-csPCa),
    fontupper=\fontsize{8pt}{9.5pt}\selectfont,
]
Review the prostate MRI images as one volumetric examination.

\smallskip
Choose the more likely diagnosis:\\

- NO\_CSPCA\\
- CSPCA

\smallskip
NO\_CSPCA means no clinically significant prostate cancer.\\
CSPCA means clinically significant prostate cancer.

\smallskip
Give one brief paragraph stating the main distinguishing evidence.

\smallskip

End with exactly one line:\\

Final Answer: NO\_CSPCA\\
or\\
Final Answer: CSPCA
\end{tcolorbox}

\vspace{0.5cm}

\begin{tcolorbox}[
    fonttitle=\footnotesize\bfseries,
    colback=gray!5!white,
    colframe=black!60,
    left=4pt,
    right=4pt,
    top=3pt,
    bottom=3pt,
    title=Liver MRI Diagnostic Prompt (HCC+ vs.\ HCC$-$),
    fontupper=\fontsize{8pt}{9.5pt}\selectfont,
]
Review the liver MRI images as one volumetric examination.

\smallskip
Choose the more likely diagnosis:\\

- HCC\_NEGATIVE\\
- HCC\_POSITIVE

\smallskip
HCC\_NEGATIVE means no hepatocellular carcinoma is present.\\
HCC\_POSITIVE means hepatocellular carcinoma is present.

\smallskip
Give one brief paragraph stating the main distinguishing evidence.

\smallskip

End with exactly one line:\\

Final Answer: HCC\_NEGATIVE\\
or\\
Final Answer: HCC\_POSITIVE
\end{tcolorbox}

\paragraph{Continuation Cues.}
For staged visual inference and text-only continuation ablations, we evaluate three cue configurations when resuming generation: a minimal hesitation cue (\texttt{Wait}), direct visual injection without any cue (\textemdash), and an explicit attention-directing cue for ablation study (\texttt{New evidence}: \textit{``Wait, these additional images may reveal something important.''}).

\subsection{Intervention Protocol}
\label{sec:supp_decoding}

\paragraph{Intervention Trigger ($\tau$).}
In standard mode (\emph{Think OFF}), decoding from the initial sequence $I_A$ pauses upon detecting the commitment prefix (\texttt{Final Answer:}); in thinking mode (\emph{Think ON}), decoding pauses upon encountering the closing reasoning tag (\texttt{</think>}). In both regimes, the commitment boundary is withheld prior to state insertion so the retained context remains uncommitted.

\paragraph{Architecture-Aware Multimodal State Extension.}
Unlike text-only test-time scaling, extending an active multimodal inference state mid-decoding without prefix replay (\Cref{eq:cache_extension}) requires maintaining representation and positional continuity across two distinct architectural state paradigms:
\begin{itemize}
\setlength\itemsep{0.3cm}
    \item \textbf{Hybrid Recurrent--Attention with 3D M-RoPE.} Qwen3.5-9B uses a 32-layer hybrid stack interleaving 24 linear-attention \emph{Gated DeltaNet} layers (maintaining recurrent state) with 8 full \emph{Gated Attention} layers (maintaining KV cache and interleaved 3D M-RoPE). Our $\operatorname{Extend}(C_\tau, B(I_B, c))$ operator executes a single incremental forward pass over $B(I_B, c)$ conditioned on $C_\tau$: advancing recurrent DeltaNet states in place, appending new visual--text KV entries in full-attention layers, computing local 3D spatio-temporal coordinates via \texttt{get\_rope\_index} over \texttt{mm\_token\_type\_ids} and \texttt{image\_grid\_thw} anchored at cache sequence length $|C_\tau|$ plus \texttt{rope\_deltas}, and immediately updating \texttt{rope\_deltas} so subsequent token decoding progresses monotonically.
    \item \textbf{ViT--MLP--LLM Embedding Injection with 1D Cache Position.} InternVL3.5-8B couples \texttt{InternViT-300M} (compressing each $448\times 448$ axial frame into 256 visual tokens) through a multilayer perceptron (MLP) projector with a \texttt{Qwen3-8B} language backbone. Visual representations extracted for $I_B$ replace \texttt{<IMG\_CONTEXT>} placeholder embeddings inside $B(I_B, c)$ prior to extending \texttt{past\_key\_values} along contiguous 1D \texttt{cache\_position} indices starting at $|C_\tau|$.
\end{itemize}
Across both architectures, the \texttt{attention\_mask} is extended across $B(I_B, c)$, and decoding verifies cache sequence-length progression without prefix replay (enforcing exact visual-token-to-feature alignment, along with pre-injection, post-injection ($|C_\tau^+| = |C_\tau| + |B(I_B, c)|$), and end-to-end cache-length assertions in InternVL3.5-8B, as well as step-wise cache-length assertions in Qwen3.5-9B).

\paragraph{Decoding Settings and Output Parsing.}
All primary experiments (\emph{Think OFF}) use deterministic greedy decoding (\texttt{do\_sample=False}) with fixed seed~42. For \emph{Think ON} ablations, each model follows its official recommended reasoning configuration to prevent repetitive looping in long chains: Qwen3.5-9B uses native thinking mode ($\text{temperature}=1.0$, $\text{top\_p}=0.95$, $\text{top\_k}=20$, $\text{presence\_penalty}=1.5$, deterministic per-case seed), whereas InternVL3.5-8B uses its official \texttt{R1\_SYSTEM\_PROMPT} \texttt{<think>...</think>} protocol ($\text{do\_sample=True}$, $\text{temperature}=0.6$, seed~42). Case-level predictions are extracted from the post-intervention text (and strictly post-\texttt{</think>} in \emph{Think ON}) via case-insensitive matching on \texttt{Final Answer: <Label>}.

\subsection{Compute Infrastructure}
\label{sec:supp_compute}

All experiments were executed in \texttt{bfloat16} precision on NVIDIA GeForce RTX~3090 GPUs (24\,GB VRAM per GPU) using PyTorch and Hugging Face \texttt{transformers} with each model's native multi-image implementation.

\begin{table*}[t]
 \caption{Marginal class-prediction distribution (\%) under \emph{Think OFF} on the balanced brain MRI cohort (true prevalence: 50.0\% GBM / 50.0\% MET). Bias denotes $(\text{Pred.\ GBM} - 50.0)$ in percentage points.}
  \centering
  \small
  \setlength{\tabcolsep}{4.2pt}
  \renewcommand{\arraystretch}{1.12}
  \begin{tabularx}{\textwidth}{l l c *{6}{>{\centering\arraybackslash}X}}
    \toprule
    & & & \multicolumn{3}{c}{\textbf{Qwen3.5-9B}} & \multicolumn{3}{c}{\textbf{InternVL3.5-8B}} \\
    \cmidrule(lr){4-6} \cmidrule(lr){7-9}
    \textbf{Condition} & \textbf{Input Flow} & \textbf{Cue} &
    \textbf{Pred.\ GBM} & \textbf{Pred.\ MET} & \textbf{Bias (pp)} &
    \textbf{Pred.\ GBM} & \textbf{Pred.\ MET} & \textbf{Bias (pp)} \\
    \midrule
    \multirow{2}{*}{Joint}
      & $(\underline{\text{T1CE}, \text{T2}})$ & \textemdash & 16.7 & 83.3 & $-$33.3 & 14.4 & 85.6 & $-$35.6 \\
      & $(\underline{\text{T2}, \text{T1CE}})$ & \textemdash &  8.9 & 91.1 & $-$41.1 & 15.6 & 84.4 & $-$34.4 \\
    \midrule
    \multirow{2}{*}{Text-Only}
      & $(\underline{\text{T1CE}, \text{T2}})$ & \texttt{Wait} & 75.6 & 24.4 & +25.6 & 36.7 & 63.3 & $-$13.3 \\
      & $(\underline{\text{T2}, \text{T1CE}})$ & \texttt{Wait} & 77.8 & 22.2 & +27.8 & 52.2 & 47.8 & +2.2 \\
    \midrule
    \cellcolor{blue!6} & \cellcolor{blue!6}$\text{T1CE} \rightarrow \text{T2}$ & \cellcolor{blue!6}\textemdash           & \cellcolor{blue!6}18.9 & \cellcolor{blue!6}81.1 & \cellcolor{blue!6}$-$31.1 & \cellcolor{blue!6}45.6 & \cellcolor{blue!6}54.4 & \cellcolor{blue!6}$-$4.4 \\
    \cellcolor{blue!6} & \cellcolor{blue!6}$\text{T1CE} \rightarrow \text{T2}$ & \cellcolor{blue!6}\texttt{Wait}      & \cellcolor{blue!6}30.0 & \cellcolor{blue!6}70.0 & \cellcolor{blue!6}$-$20.0 & \cellcolor{blue!6}46.7 & \cellcolor{blue!6}53.3 & \cellcolor{blue!6}$-$3.3 \\
    \cellcolor{blue!6} & \cellcolor{blue!6}$\text{T1CE} \rightarrow \text{T2}$ & \cellcolor{blue!6}\texttt{New evid.} & \cellcolor{blue!6}23.3 & \cellcolor{blue!6}76.7 & \cellcolor{blue!6}$-$26.7 & \cellcolor{blue!6}50.0 & \cellcolor{blue!6}50.0 & \cellcolor{blue!6}0.0 \\
    \cellcolor{blue!6} & \cellcolor{blue!6}$\text{T2} \rightarrow \text{T1CE}$ & \cellcolor{blue!6}\textemdash           & \cellcolor{blue!6}48.9 & \cellcolor{blue!6}51.1 & \cellcolor{blue!6}$-$1.1  & \cellcolor{blue!6}37.8 & \cellcolor{blue!6}62.2 & \cellcolor{blue!6}$-$12.2 \\
    \cellcolor{blue!6} & \cellcolor{blue!6}$\text{T2} \rightarrow \text{T1CE}$ & \cellcolor{blue!6}\texttt{Wait}      & \cellcolor{blue!6}45.6 & \cellcolor{blue!6}54.4 & \cellcolor{blue!6}$-$4.4  & \cellcolor{blue!6}40.0 & \cellcolor{blue!6}60.0 & \cellcolor{blue!6}$-$10.0 \\
    \cellcolor{blue!6}\multirow{-6}{*}{\textbf{Staged}}
      & \cellcolor{blue!6}$\text{T2} \rightarrow \text{T1CE}$ & \cellcolor{blue!6}\texttt{New evid.} & \cellcolor{blue!6}45.6 & \cellcolor{blue!6}54.4 & \cellcolor{blue!6}$-$4.4  & \cellcolor{blue!6}38.9 & \cellcolor{blue!6}61.1 & \cellcolor{blue!6}$-$11.1 \\
    \bottomrule
  \end{tabularx}
 
  \label{tab:supp_class_balance}
\end{table*}

\begin{table*}[t]
  \caption{Point estimates and 95\% confidence intervals (\%) for all primary \emph{Think OFF} evaluations on the brain MRI cohort.}
  \centering
  \small
  \setlength{\tabcolsep}{2.5pt}
  \renewcommand{\arraystretch}{1.08}
  \begin{tabularx}{\textwidth}{l l l *{3}{>{\hsize=0.97\hsize\centering\arraybackslash}X} >{\hsize=1.09\hsize\centering\arraybackslash}X}
    \toprule
    \textbf{Model} & \textbf{Condition} & \textbf{Input Flow (Cue)} &
    \textbf{\begin{tabular}{@{}c@{}}Bal.\ Acc.\\{[95\% CI]}\end{tabular}} &
    \textbf{\begin{tabular}{@{}c@{}}Macro-F1\\{[95\% CI]}\end{tabular}} &
    \textbf{\begin{tabular}{@{}c@{}}GBM Rec.\\{[95\% CI]}\end{tabular}} &
    \textbf{\begin{tabular}{@{}c@{}}MET Rec.\\{[95\% CI]}\end{tabular}} \\
    \midrule
    \multirow{14}{*}{Qwen3.5-9B}
      & \multirow{2}{*}{Single}
        & $\text{T1CE}$ & 64.4 [57.4, 71.7] & 60.0 [49.1, 70.3] & 31.1 [18.2, 46.6] & 97.8 [88.2, 99.9] \\
      & & $\text{T2}$   & 65.6 [55.6, 75.3] & 65.5 [55.4, 75.1] & 62.2 [46.5, 76.2] & 68.9 [53.4, 81.8] \\
      \cmidrule{2-7}
      & \multirow{2}{*}{Joint}
        & $(\underline{\text{T1CE}, \text{T2}})$ & 64.4 [57.5, 71.6] & 60.0 [49.1, 70.0] & 31.1 [18.2, 46.6] & 97.8 [88.2, 99.9] \\
      & & $(\underline{\text{T2}, \text{T1CE}})$ & 58.9 [53.6, 64.9] & 50.5 [40.2, 60.9] & 17.8 [8.0, 32.1]  & 100.0 [92.1, 100.0] \\
      \cmidrule{2-7}
      & \multirow{4}{*}{Text-Only}
        & $\text{T1CE}$ (\texttt{Wait})                    & 48.9 [38.9, 58.8] & 48.0 [37.5, 57.9] & 62.2 [46.5, 76.2] & 35.6 [21.9, 51.2] \\
      & & $\text{T2}$ (\texttt{Wait})                      & 40.0 [30.1, 49.9] & 39.5 [29.3, 49.1] & 31.1 [18.2, 46.6] & 48.9 [33.7, 64.2] \\
      & & $(\underline{\text{T1CE}, \text{T2}})$ (\texttt{Wait}) & 43.3 [34.5, 52.4] & 39.4 [29.7, 49.4] & 68.9 [53.4, 81.8] & 17.8 [8.0, 32.1]  \\
      & & $(\underline{\text{T2}, \text{T1CE}})$ (\texttt{Wait}) & 50.0 [41.2, 58.7] & 45.8 [35.4, 56.0] & 77.8 [62.9, 88.8] & 22.2 [11.2, 37.1] \\
      \cmidrule{2-7}
      & \cellcolor{blue!6} & \cellcolor{blue!6}$\text{T1CE} \rightarrow \text{T2}$ (\textemdash)        & \cellcolor{blue!6}60.0 [52.2, 67.8] & \cellcolor{blue!6}55.7 [45.1, 66.1] & \cellcolor{blue!6}28.9 [16.4, 44.3] & \cellcolor{blue!6}91.1 [78.8, 97.5] \\
      & \cellcolor{blue!6} & \cellcolor{blue!6}$\text{T1CE} \rightarrow \text{T2}$ (\texttt{Wait})      & \cellcolor{blue!6}64.4 [55.2, 73.3] & \cellcolor{blue!6}63.0 [52.3, 72.7] & \cellcolor{blue!6}44.4 [29.6, 60.0] & \cellcolor{blue!6}84.4 [70.5, 93.5] \\
      & \cellcolor{blue!6} & \cellcolor{blue!6}$\text{T1CE} \rightarrow \text{T2}$ (\texttt{New evid.}) & \cellcolor{blue!6}60.0 [51.6, 68.4] & \cellcolor{blue!6}56.9 [46.2, 67.3] & \cellcolor{blue!6}33.3 [20.0, 49.0] & \cellcolor{blue!6}86.7 [73.2, 94.9] \\
      & \cellcolor{blue!6} & \cellcolor{blue!6}$\text{T2} \rightarrow \text{T1CE}$ (\textemdash)        & \cellcolor{blue!6}\textbf{76.7} [67.6, 85.3] & \cellcolor{blue!6}\textbf{76.7} [67.5, 85.3] & \cellcolor{blue!6}75.6 [60.5, 87.1] & \cellcolor{blue!6}77.8 [62.9, 88.8] \\
      & \cellcolor{blue!6} & \cellcolor{blue!6}$\text{T2} \rightarrow \text{T1CE}$ (\texttt{Wait})      & \cellcolor{blue!6}75.6 [66.5, 84.3] & \cellcolor{blue!6}75.5 [66.2, 84.2] & \cellcolor{blue!6}71.1 [55.7, 83.6] & \cellcolor{blue!6}80.0 [65.4, 90.4] \\
      & \cellcolor{blue!6}\multirow{-6}{*}{\textbf{Staged}}
        & \cellcolor{blue!6}$\text{T2} \rightarrow \text{T1CE}$ (\texttt{New evid.}) & \cellcolor{blue!6}71.1 [61.6, 80.2] & \cellcolor{blue!6}71.1 [61.1, 80.0] & \cellcolor{blue!6}66.7 [51.0, 80.0] & \cellcolor{blue!6}75.6 [60.5, 87.1] \\
    \midrule
    \multirow{14}{*}{InternVL3.5-8B}
      & \multirow{2}{*}{Single}
        & $\text{T1CE}$ & 55.6 [50.3, 61.2] & 45.9 [36.3, 55.9] & 13.3 [5.1, 26.8]  & 97.8 [88.2, 99.9] \\
      & & $\text{T2}$   & 57.8 [48.3, 67.2] & 56.0 [45.3, 66.1] & 37.8 [23.8, 53.5] & 77.8 [62.9, 88.8] \\
      \cmidrule{2-7}
      & \multirow{2}{*}{Joint}
        & $(\underline{\text{T1CE}, \text{T2}})$ & 60.0 [53.2, 67.1] & 54.2 [43.3, 64.6] & 24.4 [12.9, 39.5] & 95.6 [84.9, 99.5] \\
      & & $(\underline{\text{T2}, \text{T1CE}})$ & 54.4 [47.2, 61.9] & 48.3 [38.4, 58.4] & 20.0 [9.6, 34.6]  & 88.9 [75.9, 96.3] \\
      \cmidrule{2-7}
      & \multirow{4}{*}{Text-Only}
        & $\text{T1CE}$ (\texttt{Wait})                    & 48.9 [39.0, 58.9] & 47.6 [37.0, 57.9] & 64.4 [48.8, 78.1] & 33.3 [20.0, 49.0] \\
      & & $\text{T2}$ (\texttt{Wait})                      & 46.7 [36.5, 57.0] & 46.6 [36.2, 56.6] & 51.1 [35.8, 66.3] & 42.2 [27.7, 57.8] \\
      & & $(\underline{\text{T1CE}, \text{T2}})$ (\texttt{Wait}) & 53.3 [43.2, 63.3] & 52.5 [41.8, 62.6] & 40.0 [25.7, 55.7] & 66.7 [51.0, 80.0] \\
      & & $(\underline{\text{T2}, \text{T1CE}})$ (\texttt{Wait}) & 48.9 [38.5, 59.1] & 48.9 [38.3, 58.8] & 51.1 [35.8, 66.3] & 46.7 [31.7, 62.1] \\
      \cmidrule{2-7}
      & \cellcolor{blue!6} & \cellcolor{blue!6}$\text{T1CE} \rightarrow \text{T2}$ (\textemdash)        & \cellcolor{blue!6}64.4 [54.4, 74.1] & \cellcolor{blue!6}64.4 [54.2, 73.7] & \cellcolor{blue!6}60.0 [44.3, 74.3] & \cellcolor{blue!6}68.9 [53.4, 81.8] \\
      & \cellcolor{blue!6} & \cellcolor{blue!6}$\text{T1CE} \rightarrow \text{T2}$ (\texttt{Wait})      & \cellcolor{blue!6}65.6 [55.6, 75.4] & \cellcolor{blue!6}65.5 [55.4, 75.1] & \cellcolor{blue!6}62.2 [46.5, 76.2] & \cellcolor{blue!6}68.9 [53.4, 81.8] \\
      & \cellcolor{blue!6} & \cellcolor{blue!6}$\text{T1CE} \rightarrow \text{T2}$ (\texttt{New evid.}) & \cellcolor{blue!6}66.7 [56.7, 76.3] & \cellcolor{blue!6}66.7 [56.5, 76.2] & \cellcolor{blue!6}66.7 [51.0, 80.0] & \cellcolor{blue!6}66.7 [51.0, 80.0] \\
      & \cellcolor{blue!6} & \cellcolor{blue!6}$\text{T2} \rightarrow \text{T1CE}$ (\textemdash)        & \cellcolor{blue!6}67.8 [58.3, 77.0] & \cellcolor{blue!6}67.3 [57.0, 76.6] & \cellcolor{blue!6}55.6 [40.0, 70.4] & \cellcolor{blue!6}80.0 [65.4, 90.4] \\
      & \cellcolor{blue!6} & \cellcolor{blue!6}$\text{T2} \rightarrow \text{T1CE}$ (\texttt{Wait})      & \cellcolor{blue!6}65.6 [56.0, 75.0] & \cellcolor{blue!6}65.2 [55.1, 74.5] & \cellcolor{blue!6}55.6 [40.0, 70.4] & \cellcolor{blue!6}75.6 [60.5, 87.1] \\
      & \cellcolor{blue!6}\multirow{-6}{*}{\textbf{Staged}}
        & \cellcolor{blue!6}$\text{T2} \rightarrow \text{T1CE}$ (\texttt{New evid.}) & \cellcolor{blue!6}\textbf{68.9} [59.3, 77.9] & \cellcolor{blue!6}\textbf{68.5} [58.4, 77.7] & \cellcolor{blue!6}57.8 [42.2, 72.3] & \cellcolor{blue!6}80.0 [65.4, 90.4] \\
    \bottomrule
  \end{tabularx}
  \label{tab:supp_ci_off}
\end{table*}

\section{Extended Ablations}
\label{sec:supp_ablations}

This section provides targeted ablations complementing \Cref{tab:primary_comparison,tab:text_vs_staged,tab:thinking_ablation} of the main paper. We isolate three factors relevant to test-time scaling reliability: sensitivity to verbal continuation cues, the effect of built-in thinking across both sequence orders, and how joint, text-only, and staged inference alter marginal class-prediction balance. Every evaluation across these brain MRI conditions yielded a valid diagnostic prediction with zero parse failures.

\subsection{Cue-Framing Sensitivity}
\label{sec:supp_cue_ablation}

A key question for staged inference is whether gains arise from explicit prompt engineering around the \texttt{Wait} cue, or whether stronger instructions emphasizing the newly arriving images could rescue unfavorable sequence orders. \Cref{tab:supp_cue_ablation} compares three cue variants under thinking-disabled inference: cue-free injection (\textemdash), minimal hesitation (\texttt{Wait}), and an explicit attention-directing cue (\texttt{New evidence}: \textit{``Wait, these additional images may reveal something important.''}).

Across both architectures, all three cue settings along $\text{T2} \rightarrow \text{T1CE}$ substantially improve over the order-matched $(\underline{\text{T2}, \text{T1CE}})$ baseline, confirming that visual staging—rather than specific wording—drives the gain. Notably, adding persuasive framing (\texttt{New evidence}) offers no consistent benefit over cue-free or minimal injection. On Qwen3.5-9B, \texttt{New evidence} slightly degrades $\text{T2} \rightarrow \text{T1CE}$ performance relative to cue-free injection (inducing 21 prediction flips, of which 13 are regressions vs.\ 8 fixes) while lengthening post-injection explanations. One plausible explanation is that stronger instruction-tuned models interpret suggestive phrasing as an implicit directive to revise an already calibrated initial T2 assessment. Conversely, InternVL3.5-8B produces much shorter initial prefixes ($\sim$80--84 tokens vs.\ 172--174 on Qwen3.5-9B) and exhibits slightly higher stability and accuracy under \texttt{New evidence}. This prefix-length difference also aligns with the order asymmetry observed under $\text{T1CE} \rightarrow \text{T2}$: Qwen3.5-9B commits a longer initial rationale to its KV cache when starting on T1CE, resisting late T2 updates across all cues, whereas InternVL3.5-8B's shorter prefix permits consistent improvement regardless of cue wording.

\subsection{Built-In Thinking}
\label{sec:supp_thinking_orders}

To verify that the findings in \Cref{tab:thinking_ablation} of the main paper are not specific to the $\text{T2}$-first pathway or the \texttt{Wait} cue, \Cref{tab:supp_thinking_full} extends the \emph{Think OFF} vs.\ \emph{Think ON} comparison across both sequence directions and cue settings.

On Qwen3.5-9B, enabling thinking never recovers the gains of concise staged inference under any order or cue, keeping balanced accuracy trapped near joint baselines while inflating token consumption by an order of magnitude. On InternVL3.5-8B, \emph{Think ON} reveals a sharp contrast between pathways that aligns with pre-injection chain length: along $\text{T2} \rightarrow \text{T1CE}$, pre-injection reasoning expands to nearly 2,650--2,900 tokens and limits or degrades post-injection revision (yielding no gain over \emph{Think OFF} without a cue and dropping to 58.9\% with \texttt{Wait}); along $\text{T1CE} \rightarrow \text{T2}$, pre-injection chains stay roughly half as long, allowing \texttt{Wait}-cued \emph{Think ON} to reach 71.1\% balanced accuracy—though at roughly 9$\times$ the latency and 12$\times$ the token cost of \emph{Think OFF} staging, and with renewed drop-off when the cue is omitted. Consistent with the visual-attention decay observed in \Cref{fig:visual_attention} of the main paper, prolonged internal deliberation tends to weaken cross-modal revision unless pre-injection chains remain tightly bounded. Concise \emph{Think OFF} staging avoids this fragility while providing a substantially more favorable accuracy--compute trade-off across both backbones.

\subsection{Class-Prediction Balance}
\label{sec:supp_class_balance}

To clarify why text-only continuation reduces accuracy in \Cref{tab:text_vs_staged} of the main paper rather than merely leaving predictions unchanged, \Cref{tab:supp_class_balance} reports the marginal class-prediction proportions under \emph{Think OFF} on this balanced (50\% GBM / 50\% MET) cohort.

Joint presentation suffers severe modality dominance toward MET, assigning over 83\% of cases to MET on both models and missing most GBM cases. When interrupted with text-only \texttt{Wait} without new visual input, models do not maintain calibrated confidence on their cached images; instead, they treat the hesitation prompt as a conversational cue to overturn their initial decision, swinging sharply toward GBM over-prediction while lowering overall accuracy. Staged $\text{T2} \rightarrow \text{T1CE}$ presentation avoids both extremes—bringing Qwen3.5-9B close to balanced class parity and mitigating InternVL3.5-8B's MET skew—confirming that introducing complementary visual evidence supports calibrated cross-sequence adjustment rather than ungrounded label reversal.

\begin{table*}[t]
  \caption{Point estimates and 95\% confidence intervals (\%) for dual-sequence \emph{Think ON} ablations on the brain MRI cohort.}
  \centering
  \small
  \setlength{\tabcolsep}{4pt}
  \renewcommand{\arraystretch}{1.10}
  \begin{tabularx}{\textwidth}{l l l *{4}{>{\centering\arraybackslash}X}}
    \toprule
    \textbf{Model} & \textbf{Condition} & \textbf{Input Flow (Cue)} &
    \textbf{Bal.\ Acc.\ [95\% CI]} & \textbf{Macro-F1 [95\% CI]} &
    \textbf{GBM Rec.\ [95\% CI]} & \textbf{MET Rec.\ [95\% CI]} \\
    \midrule
    \multirow{6}{*}{Qwen3.5-9B}
      & \multirow{2}{*}{Joint}
        & $(\underline{\text{T1CE}, \text{T2}})$ & 62.2 [54.1, 70.4] & 58.8 [47.6, 68.9] & 33.3 [20.0, 49.0] & 91.1 [78.8, 97.5] \\
      & & $(\underline{\text{T2}, \text{T1CE}})$ & 58.9 [52.2, 65.9] & 52.5 [41.7, 62.7] & 22.2 [11.2, 37.1] & 95.6 [84.9, 99.5] \\
      \cmidrule{2-7}
      & \cellcolor{blue!6} & \cellcolor{blue!6}$\text{T1CE} \rightarrow \text{T2}$ (\textemdash)   & \cellcolor{blue!6}61.1 [54.3, 68.3] & \cellcolor{blue!6}55.9 [45.3, 66.1] & \cellcolor{blue!6}26.7 [14.6, 41.9] & \cellcolor{blue!6}95.6 [84.9, 99.5] \\
      & \cellcolor{blue!6} & \cellcolor{blue!6}$\text{T1CE} \rightarrow \text{T2}$ (\texttt{Wait}) & \cellcolor{blue!6}\textbf{64.4} [56.7, 72.3] & \cellcolor{blue!6}\textbf{61.2} [50.6, 71.2] & \cellcolor{blue!6}35.6 [21.9, 51.2] & \cellcolor{blue!6}93.3 [81.7, 98.6] \\
      & \cellcolor{blue!6} & \cellcolor{blue!6}$\text{T2} \rightarrow \text{T1CE}$ (\textemdash)   & \cellcolor{blue!6}57.8 [51.4, 64.6] & \cellcolor{blue!6}50.7 [40.3, 61.1] & \cellcolor{blue!6}20.0 [9.6, 34.6]  & \cellcolor{blue!6}95.6 [84.9, 99.5] \\
      & \cellcolor{blue!6}\multirow{-4}{*}{\textbf{Staged}}
        & \cellcolor{blue!6}$\text{T2} \rightarrow \text{T1CE}$ (\texttt{Wait}) & \cellcolor{blue!6}60.0 [52.7, 67.8] & \cellcolor{blue!6}55.0 [44.2, 65.6] & \cellcolor{blue!6}26.7 [14.6, 41.9] & \cellcolor{blue!6}93.3 [81.7, 98.6] \\
    \midrule
    \multirow{6}{*}{InternVL3.5-8B}
      & \multirow{2}{*}{Joint}
        & $(\underline{\text{T1CE}, \text{T2}})$ & 64.4 [54.6, 74.0] & 64.2 [54.0, 73.7] & 55.6 [40.0, 70.4] & 73.3 [58.1, 85.4] \\
      & & $(\underline{\text{T2}, \text{T1CE}})$ & 65.6 [55.8, 75.4] & 65.3 [55.1, 75.2] & 57.8 [42.2, 72.3] & 73.3 [58.1, 85.4] \\
      \cmidrule{2-7}
      & \cellcolor{blue!6} & \cellcolor{blue!6}$\text{T1CE} \rightarrow \text{T2}$ (\textemdash)   & \cellcolor{blue!6}66.7 [57.0, 76.0] & \cellcolor{blue!6}66.4 [56.4, 75.6] & \cellcolor{blue!6}57.8 [42.2, 72.3] & \cellcolor{blue!6}75.6 [60.5, 87.1] \\
      & \cellcolor{blue!6} & \cellcolor{blue!6}$\text{T1CE} \rightarrow \text{T2}$ (\texttt{Wait}) & \cellcolor{blue!6}\textbf{71.1} [61.9, 80.1] & \cellcolor{blue!6}\textbf{70.8} [60.9, 79.9] & \cellcolor{blue!6}60.0 [44.3, 74.3] & \cellcolor{blue!6}82.2 [67.9, 92.0] \\
      & \cellcolor{blue!6} & \cellcolor{blue!6}$\text{T2} \rightarrow \text{T1CE}$ (\textemdash)   & \cellcolor{blue!6}67.8 [57.8, 77.6] & \cellcolor{blue!6}67.8 [57.6, 77.5] & \cellcolor{blue!6}66.7 [51.0, 80.0] & \cellcolor{blue!6}68.9 [53.4, 81.8] \\
      & \cellcolor{blue!6}\multirow{-4}{*}{\textbf{Staged}}
        & \cellcolor{blue!6}$\text{T2} \rightarrow \text{T1CE}$ (\texttt{Wait}) & \cellcolor{blue!6}58.9 [48.9, 68.9] & \cellcolor{blue!6}58.8 [48.5, 68.8] & \cellcolor{blue!6}64.4 [48.8, 78.1] & \cellcolor{blue!6}53.3 [37.9, 68.3] \\
    \bottomrule
  \end{tabularx}

  \label{tab:supp_ci_on}
\end{table*}

\begin{table*}[t]
  \caption{Paired prediction transitions and paired bootstrap 95\% CIs relative to order-matched joint baselines across dual-sequence \emph{Think OFF} evaluations ($\text{W}$: wrong; $\text{C}$: correct; $\Delta\text{BA}$/$\Delta\text{F1}$: changes in Balanced Accuracy and Macro-F1). Best values per model are in \textbf{bold}.}
  \centering
  \footnotesize
  \setlength{\tabcolsep}{3.8pt}
  \renewcommand{\arraystretch}{1.12}
  \begin{tabularx}{\textwidth}{l X c c c c c c c}
    \toprule
    \textbf{Condition} & \textbf{Comparison vs.\ Joint} &
    \textbf{Flip\%} & \textbf{W$\rightarrow$C} & \textbf{C$\rightarrow$W} & \textbf{Net} &
    \textbf{C$\rightarrow$C / W$\rightarrow$W} & \textbf{$\Delta$BA [95\% CI]} & \textbf{$\Delta$F1 [95\% CI]} \\
    \midrule
    \multicolumn{9}{c}{\textbf{Qwen3.5-9B}} \\
    \midrule
    \multirow{2}{*}{Text-Only}
      & $(\underline{\text{T1CE}, \text{T2}}) + \texttt{Wait}$ & 72.2 & 23 & 42 & $-$19 & 16 / 9 & $-$21.1 [$-$33.3, $-$9.1] & $-$20.6 [$-$36.1, $-$4.9] \\
      & $(\underline{\text{T2}, \text{T1CE}}) + \texttt{Wait}$ & 75.6 & 30 & 38 & $-$8  & 15 / 7 & $-$8.9 [$-$20.2, +2.0]   & $-$4.7 [$-$20.8, +11.1] \\
    \midrule
    \cellcolor{blue!6} & \cellcolor{blue!6}$\text{T1CE} \rightarrow \text{T2}$ (\textemdash)        & \cellcolor{blue!6}20.0 & \cellcolor{blue!6}7  & \cellcolor{blue!6}11 & \cellcolor{blue!6}$-$4 & \cellcolor{blue!6}47 / 25 & \cellcolor{blue!6}$-$4.4 [$-$13.9, +4.6] & \cellcolor{blue!6}$-$4.3 [$-$16.3, +7.3] \\
    \cellcolor{blue!6} & \cellcolor{blue!6}$\text{T1CE} \rightarrow \text{T2}$ (\texttt{Wait})      & \cellcolor{blue!6}26.7 & \cellcolor{blue!6}12 & \cellcolor{blue!6}12 & \cellcolor{blue!6}0   & \cellcolor{blue!6}46 / 20 & \cellcolor{blue!6}+0.0 [$-$10.8, +10.4] & \cellcolor{blue!6}+3.0 [$-$9.5, +15.6] \\
    \cellcolor{blue!6} & \cellcolor{blue!6}$\text{T1CE} \rightarrow \text{T2}$ (\texttt{New evid.}) & \cellcolor{blue!6}24.4 & \cellcolor{blue!6}9  & \cellcolor{blue!6}13 & \cellcolor{blue!6}$-$4 & \cellcolor{blue!6}45 / 23 & \cellcolor{blue!6}$-$4.4 [$-$14.7, +5.4] & \cellcolor{blue!6}$-$3.1 [$-$15.7, +9.4] \\
    \cellcolor{blue!6} & \cellcolor{blue!6}$\text{T2} \rightarrow \text{T1CE}$ (\textemdash)        & \cellcolor{blue!6}40.0 & \cellcolor{blue!6}26 & \cellcolor{blue!6}10 & \cellcolor{blue!6}\textbf{+16} & \cellcolor{blue!6}43 / 11 & \cellcolor{blue!6}\textbf{+17.8} [+8.2, +27.1] & \cellcolor{blue!6}\textbf{+26.1} [+13.8, +38.6] \\
    \cellcolor{blue!6} & \cellcolor{blue!6}$\text{T2} \rightarrow \text{T1CE}$ (\texttt{Wait})      & \cellcolor{blue!6}38.9 & \cellcolor{blue!6}25 & \cellcolor{blue!6}10 & \cellcolor{blue!6}+15 & \cellcolor{blue!6}43 / 12 & \cellcolor{blue!6}+16.7 [+6.6, +26.4] & \cellcolor{blue!6}+25.0 [+12.2, +37.6] \\
    \cellcolor{blue!6}\multirow{-6}{*}{\textbf{Staged}}
      & \cellcolor{blue!6}$\text{T2} \rightarrow \text{T1CE}$ (\texttt{New evid.}) & \cellcolor{blue!6}36.7 & \cellcolor{blue!6}22 & \cellcolor{blue!6}11 & \cellcolor{blue!6}+11 & \cellcolor{blue!6}42 / 15 & \cellcolor{blue!6}+12.2 [+2.5, +21.9] & \cellcolor{blue!6}+20.5 [+8.6, +32.6] \\
    \midrule
    \multicolumn{9}{c}{\textbf{InternVL3.5-8B}} \\
    \midrule
    \multirow{2}{*}{Text-Only}
      & $(\underline{\text{T1CE}, \text{T2}}) + \texttt{Wait}$ & 48.9 & 19 & 25 & $-$6 & 29 / 17 & $-$6.7 [$-$20.8, +6.9] & $-$1.7 [$-$18.8, +14.5] \\
      & $(\underline{\text{T2}, \text{T1CE}}) + \texttt{Wait}$ & 63.3 & 26 & 31 & $-$5 & 18 / 15 & $-$5.6 [$-$20.2, +8.7] & +0.6 [$-$16.2, +16.5] \\
    \midrule
    \cellcolor{blue!6} & \cellcolor{blue!6}$\text{T1CE} \rightarrow \text{T2}$ (\textemdash)        & \cellcolor{blue!6}35.6 & \cellcolor{blue!6}18 & \cellcolor{blue!6}14 & \cellcolor{blue!6}+4  & \cellcolor{blue!6}40 / 18 & \cellcolor{blue!6}+4.4 [$-$6.2, +14.9] & \cellcolor{blue!6}+10.2 [$-$2.6, +23.1] \\
    \cellcolor{blue!6} & \cellcolor{blue!6}$\text{T1CE} \rightarrow \text{T2}$ (\texttt{Wait})      & \cellcolor{blue!6}36.7 & \cellcolor{blue!6}19 & \cellcolor{blue!6}14 & \cellcolor{blue!6}+5  & \cellcolor{blue!6}40 / 17 & \cellcolor{blue!6}+5.6 [$-$5.2, +16.1] & \cellcolor{blue!6}+11.3 [$-$1.8, +24.4] \\
    \cellcolor{blue!6} & \cellcolor{blue!6}$\text{T1CE} \rightarrow \text{T2}$ (\texttt{New evid.}) & \cellcolor{blue!6}40.0 & \cellcolor{blue!6}21 & \cellcolor{blue!6}15 & \cellcolor{blue!6}+6  & \cellcolor{blue!6}39 / 15 & \cellcolor{blue!6}+6.7 [$-$4.4, +17.4] & \cellcolor{blue!6}+12.5 [$-$0.9, +25.8] \\
    \cellcolor{blue!6} & \cellcolor{blue!6}$\text{T2} \rightarrow \text{T1CE}$ (\textemdash)        & \cellcolor{blue!6}28.9 & \cellcolor{blue!6}19 & \cellcolor{blue!6}7  & \cellcolor{blue!6}+12 & \cellcolor{blue!6}42 / 22 & \cellcolor{blue!6}+13.3 [+3.4, +23.3] & \cellcolor{blue!6}+19.0 [+8.1, +29.9] \\
    \cellcolor{blue!6} & \cellcolor{blue!6}$\text{T2} \rightarrow \text{T1CE}$ (\texttt{Wait})      & \cellcolor{blue!6}31.1 & \cellcolor{blue!6}19 & \cellcolor{blue!6}9  & \cellcolor{blue!6}+10 & \cellcolor{blue!6}40 / 22 & \cellcolor{blue!6}+11.1 [+1.1, +21.3] & \cellcolor{blue!6}+16.9 [+5.8, +28.1] \\
    \cellcolor{blue!6}\multirow{-6}{*}{\textbf{Staged}}
      & \cellcolor{blue!6}$\text{T2} \rightarrow \text{T1CE}$ (\texttt{New evid.}) & \cellcolor{blue!6}32.2 & \cellcolor{blue!6}21 & \cellcolor{blue!6}8  & \cellcolor{blue!6}\textbf{+13} & \cellcolor{blue!6}41 / 20 & \cellcolor{blue!6}\textbf{+14.4} [+4.2, +24.7] & \cellcolor{blue!6}\textbf{+20.2} [+8.8, +31.4] \\
    \bottomrule
  \end{tabularx}

  \label{tab:supp_paired_full}
\end{table*}

\begin{figure*}[t]
    \centering
    \includegraphics[width=1\textwidth]{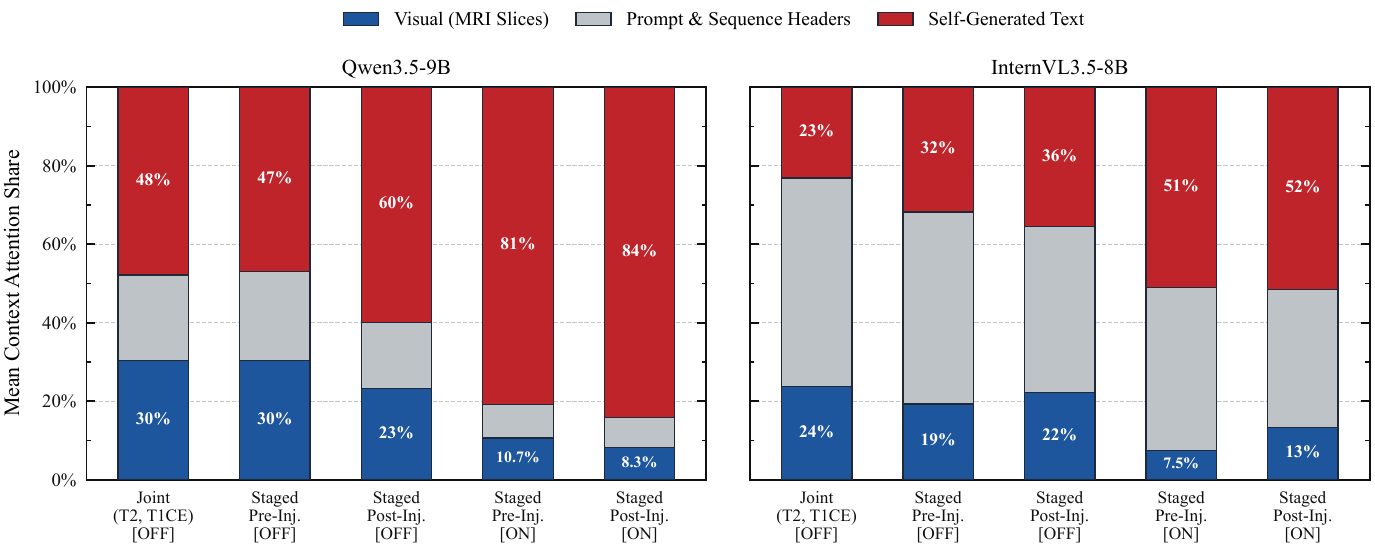}
    \caption{Mean context attention allocation across visual MRI tokens, prompt/sequence headers, and self-generated assistant text along the $\text{T2}$-first pathway. Enabling built-in thinking shifts the attention budget heavily toward self-generated reasoning tokens in both architectures.}
    \label{fig:supp_attention_budget}
\end{figure*}

\section{Statistical Analyses}
\label{sec:supp_stats}

This section provides 95\% confidence intervals (CIs), paired hypothesis tests, and complete case-level transition breakdowns on the primary brain MRI cohort.

\paragraph{Statistical Protocol.}
Single-condition 95\% CIs are computed using 10,000 patient-level bootstrap resamples for balanced accuracy and Macro-F1, and exact Clopper--Pearson binomial intervals ($n=45$ per class) for per-class recall. For paired comparisons across the same 90 patients, we report paired differences in Balanced Accuracy ($\Delta\text{BA}$) and Macro-F1 ($\Delta\text{F1}$) with 10,000-resample paired bootstrap 95\% CIs, alongside exact two-sided McNemar $p$-values ($p_{\text{McN}}$) on case-level correctness and 10,000-swap paired permutation $p$-values ($p_{\text{F1}}$) on Macro-F1. Reporting $p_{\text{F1}}$ alongside $p_{\text{McN}}$ is important because unweighted McNemar treats extreme single-class collapse (e.g., near-100\% MET recall with severe GBM under-detection) identically to a balanced classifier of equal raw accuracy, whereas Macro-F1 directly evaluates class-balanced diagnostic performance.

\subsection{Confidence Intervals}
\label{sec:supp_ci}

\Cref{tab:supp_ci_off} reports 95\% CIs for all primary thinking-disabled configurations (\Cref{tab:primary_comparison,tab:text_vs_staged} of the main paper and the cue ablation), while \Cref{tab:supp_ci_on} reports CIs for the dual-sequence thinking-enabled ablations.

With $N=90$ patients, marginal single-condition intervals naturally reflect inter-patient variability; accordingly, overlapping marginal intervals between two conditions do not imply a non-significant paired difference, since shared patient-level difficulty is removed only in paired comparisons (\Cref{sec:supp_paired_transitions}). Structurally, \Cref{tab:supp_ci_off} shows two consistent patterns under \emph{Think OFF}: text-only continuation widens uncertainty around chance-level accuracy while inverting class recall bounds, whereas staged $\text{T2} \rightarrow \text{T1CE}$ shifts both GBM and MET recall intervals upward into well-balanced operating ranges without relying on single-class skew.

\begin{figure*}[t]
    \centering
    \includegraphics[width=0.92\textwidth]{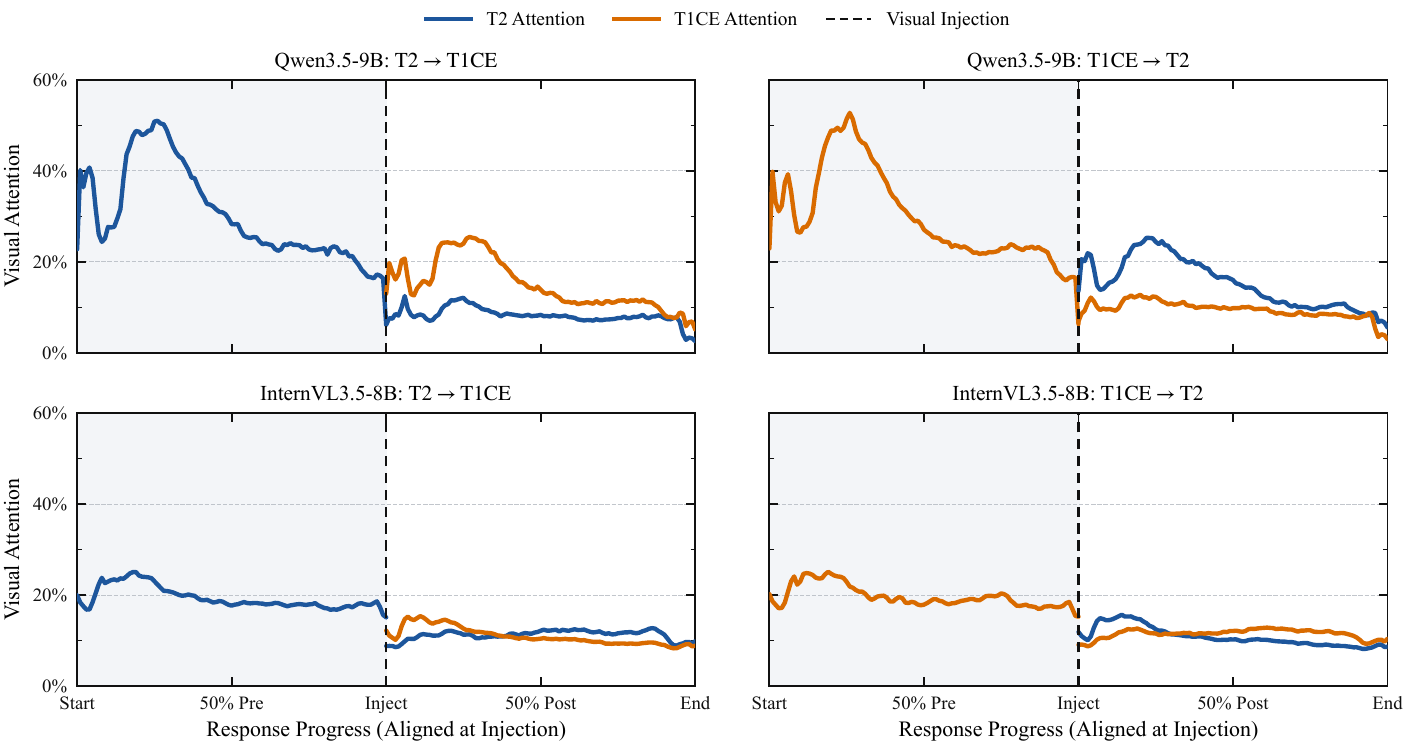}
    \caption{Visual attention between $\text{T2}$ and $\text{T1CE}$ aligned at the injection point across both staging directions (\emph{Think OFF}). Shaded regions denote pre-injection prefix decoding on the initial sequence alone; dashed lines mark complementary visual injection.}
    \label{fig:supp_sequence_flow}
\end{figure*}

\begin{figure}[t]
    \centering
    \includegraphics[width=1\columnwidth]{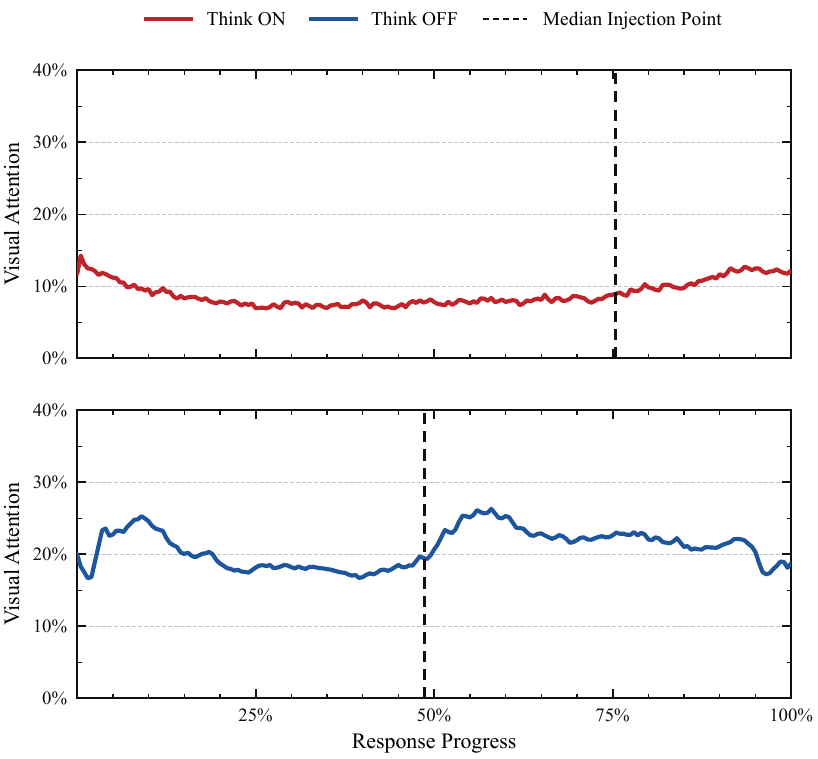}
    \caption{Visual attention share across normalized response progress for InternVL3.5-8B along the staged $\text{T2} \rightarrow \text{T1CE}$ pathway (\emph{Think ON} vs.\ \emph{Think OFF}), complementing \protect\Cref{fig:visual_attention} of the main paper.}
    \label{fig:supp_internvl_think}
\end{figure}

\begin{figure*}[t]
    \centering
    \includegraphics[width=0.92\textwidth]{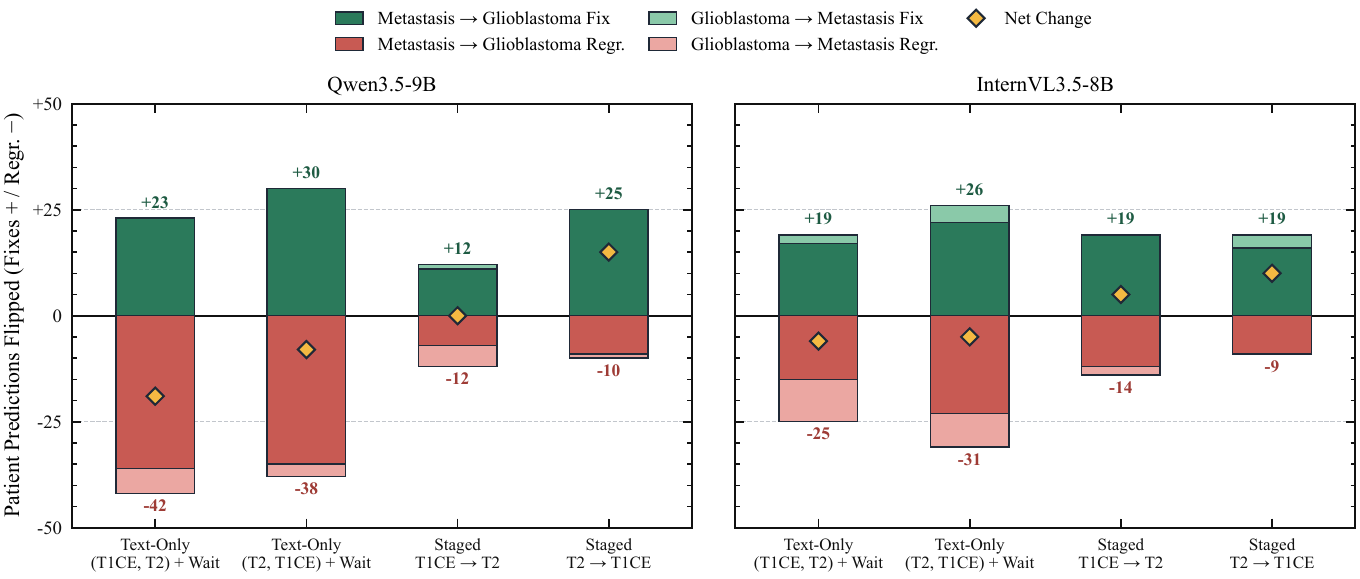}
    \caption{Case-level prediction transition directions relative to order-matched joint baselines on the full brain MRI cohort (\emph{Think OFF}, \texttt{Wait} cue). Diamonds mark net accuracy change (constructive fixes minus regressions).}
    \label{fig:supp_flip_anatomy}
\end{figure*}

\begin{figure}[t]
    \centering
    \includegraphics[width=\columnwidth]{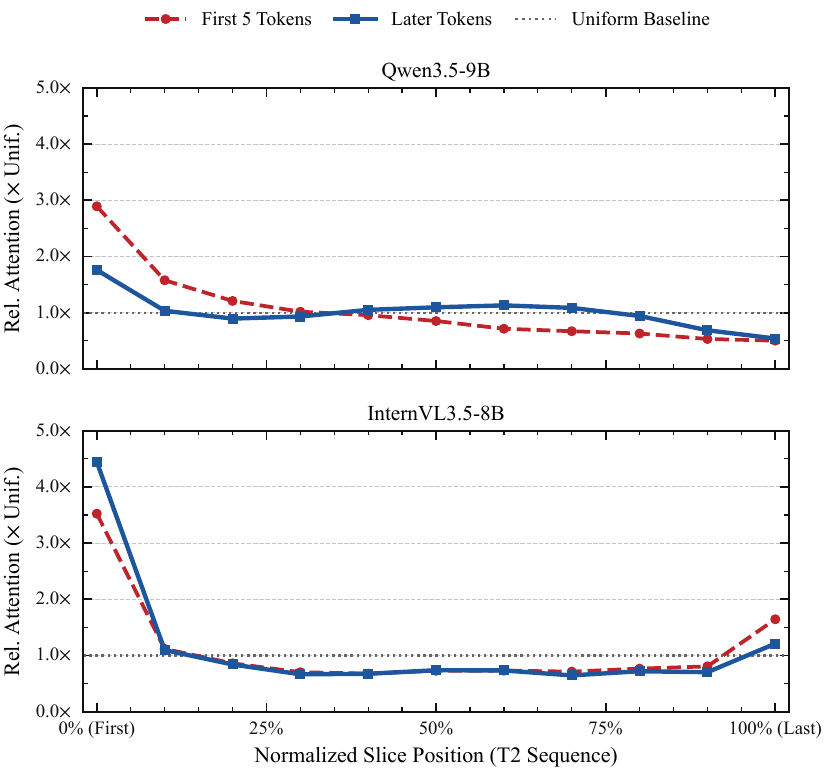}
    \caption{Relative slice attention across normalized $\text{T2}$ slice positions ($1.0\times$ denotes uniform allocation across slices) comparing the first five generated tokens against later decoding steps.}
    \label{fig:supp_boundary_sinks}
\end{figure}

\begin{figure*}[t]
    \centering
    \includegraphics[width=0.92\textwidth]{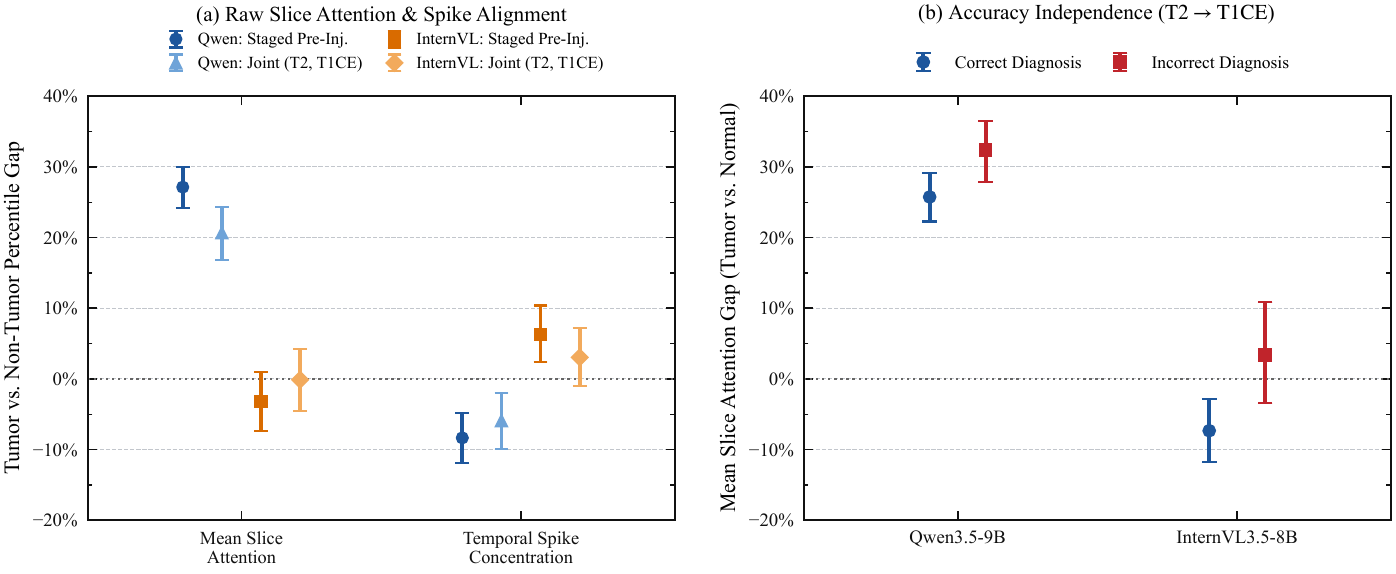}
    \caption{Exploratory $\text{T2}$ slice-level separation and case-level outcome audit (mean percentile gap with 95\% bootstrap CIs). (a) Tumor-versus-normal slice gap for time-averaged slice attention and temporal spike concentration. (b) Mean slice attention gap stratified by case-level diagnostic correctness under $\text{T2} \rightarrow \text{T1CE}$.}
    \label{fig:supp_slice_pitfalls}
\end{figure*}

\begin{figure}[t]
    \centering
    \includegraphics[width=\columnwidth]{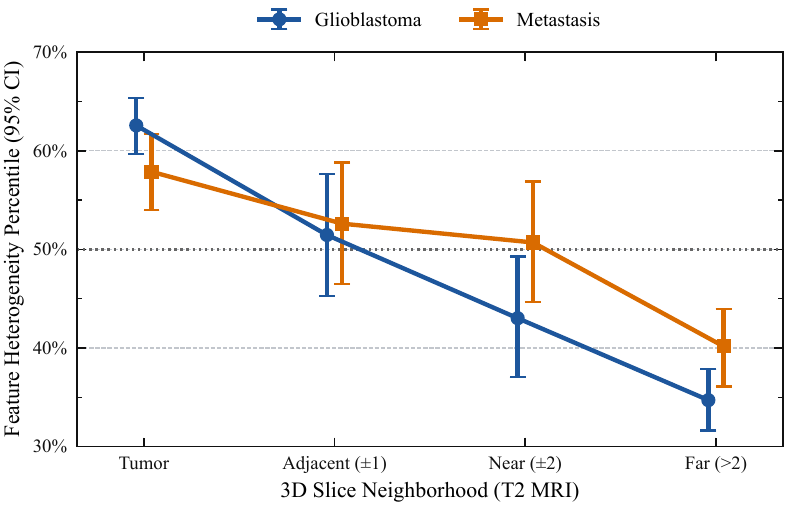}
    \caption{Mid-depth within-slice visual feature heterogeneity across $\text{T2}$ 3D slice neighborhoods on the brain MRI cohort (95\% patient-bootstrap CIs).}
    \label{fig:supp_feature_heterogeneity}
\end{figure}

\subsection{Paired Transitions and Hypothesis Tests}
\label{sec:supp_paired_transitions}

\Cref{tab:supp_paired_full} expands \Cref{tab:paired_vs_joint} of the main manuscript across all primary thinking-disabled dual-sequence conditions relative to order-matched joint baselines, reporting prediction turnover ($\text{Flip\%}$), full case-level transition counts, and paired bootstrap 95\% CIs for $\Delta\text{BA}$ and $\Delta\text{F1}$.

Together with paired significance testing, these results support three core conclusions:
\begin{itemize}
    \item \textbf{Selective revision versus ungrounded churn.} Prompting \texttt{Wait} without new visual evidence destabilizes predictions across both single-sequence and joint baselines, causing roughly half to three-quarters of patients to flip labels and yielding consistent net accuracy losses (worst on single-sequence $\text{T2}$). Staged visual injection sharply curtails this turnover and, along $\text{T2} \rightarrow \text{T1CE}$, reverses the balance so constructive fixes outnumber regressions by at least two-to-one under every cue setting. Class-stratified inspection confirms that these gains stem overwhelmingly from recovering GBM previously masked under joint $\text{T1CE}$ dominance ($p < 0.0001$ across cues in both models), while the modest remaining regressions show that staging shifts the revision balance favorably rather than making updates monotonic.
    \item \textbf{Significance over order-matched and stronger joint baselines.} Along $\text{T2} \rightarrow \text{T1CE}$, every cue setting achieves paired 95\% CIs strictly excluding zero for both $\Delta\text{BA}$ and $\Delta\text{F1}$ across both architectures (\Cref{tab:supp_paired_full}). Paired permutation tests confirm significant Macro-F1 gains in both models ($p_{\text{F1}} \le 0.002$ on Qwen3.5-9B; $p_{\text{F1}} \le 0.007$ on InternVL3.5-8B), backed by exact McNemar significance ($p_{\text{McN}} \le 0.017$ on Qwen3.5-9B for cue-free and \texttt{Wait}; $p_{\text{McN}} \le 0.029$ on InternVL3.5-8B for cue-free and \texttt{New evidence}). Notably, on Qwen3.5-9B---where $(\underline{\text{T1CE}, \text{T2}})$ is the stronger joint order---$\text{T2} \rightarrow \text{T1CE}$ also significantly outperforms that stronger joint baseline (paired 95\% CIs strictly above zero and $p_{\text{F1}} < 0.02$ for cue-free and \texttt{Wait}), verifying that staged gains are not an artifact of a weak ordering comparator.
    \item \textbf{Advantage over text-only continuation and built-in thinking.} Staged $\text{T2} \rightarrow \text{T1CE}$ (\texttt{Wait}) significantly outperforms order-matched text-only continuation across both backbones ($p_{\text{McN}} < 0.001$ on Qwen3.5-9B; $p_{\text{McN}} < 0.04$ on InternVL3.5-8B), confirming that complementary visual evidence rather than hesitation prompting drives diagnostic improvement. Relative to built-in thinking (\Cref{tab:thinking_ablation} of the main paper), concise \emph{Think OFF} $\text{T2} \rightarrow \text{T1CE}$ significantly outperforms both joint and staged \emph{Think ON} on Qwen3.5-9B ($p_{\text{McN}} \le 0.020$, $p_{\text{F1}} \le 0.002$), and matches InternVL3.5-8B joint \emph{Think ON} ($p = 1.00$) at roughly an order-of-magnitude lower compute cost.
\end{itemize}

\paragraph{External Cohort Completeness.}
Across the external cohorts without slice subsampling or lesion localization, InternVL3.5-8B exhibits weak baseline sensitivity overall, performing near chance on both PI-CAI and OpenSwissHCC. All external evaluations completed with valid predictions across the full cohorts except for the single staged InternVL3.5-8B run on OpenSwissHCC, in which three examinations did not yield a valid output; \Cref{tab:external_eval} of the main paper therefore reports metrics over the completed valid cases for that condition.

\section{Mechanistic and Exploratory Audit}
\label{sec:supp_mechanistic}

To elucidate why staged visual presentation succeeds where prolonged textual reasoning and ungrounded continuation fail, we audit context attention allocation, response-progress sequence flow, case-level revision anatomy, and exploratory slice-level signals across the brain MRI cohort. All mechanistic measurements keep model parameters frozen and mirror the exact decoding configurations of the main evaluation; for Qwen3.5-9B's hybrid stack, token-level attention shares are extracted across its eight full Gated Attention layers alongside InternVL3.5-8B's dense self-attention layers.

\subsection{Attention Grounding, Self-Conditioning, and Sequence Flow}
\label{sec:supp_attention_budget}

A central question raised by our behavioral ablations is why built-in thinking (\emph{Think ON}) degrades staged revision despite generating substantially more tokens. Partitioning the context attention budget in \Cref{fig:supp_attention_budget} clarifies the underlying mechanism. Under concise \emph{Think OFF} execution, both joint and staged inference maintain strong visual grounding before and after complementary evidence arrives. 

By contrast, enabling \emph{Think ON} induces pronounced \emph{self-conditioning}: as long reasoning chains accumulate in the KV cache, attention shifts disproportionately onto the model's own generated tokens, crowding out visual slices. This self-conditioning bottleneck is markedly stronger in Qwen3.5-9B---where self-generated text dominates the context budget and compresses visual share throughout both phases---aligning with its steeper accuracy drop under staged \emph{Think ON} relative to InternVL3.5-8B (\Cref{tab:thinking_ablation} of the main paper).

\Cref{fig:supp_internvl_think} confirms the temporal dynamics of this visual-grounding erosion on InternVL3.5-8B along the $\text{T2} \rightarrow \text{T1CE}$ pathway, directly complementing \Cref{fig:visual_attention} of the main paper. Under \emph{Think ON}, visual attention decays rapidly early in generation onto a low plateau and shows only a muted uptick after $\text{T1CE}$ injection. Under \emph{Think OFF}, by contrast, InternVL3.5-8B sustains higher visual engagement throughout prefix formation and exhibits an immediate, sustained rebound upon complementary $\text{T1CE}$ arrival. Concise inference thus co-occurs with stronger visual grounding as newly introduced evidence arrives.

To understand how staging resolves joint modality interference---and why sequence order governs performance on Qwen3.5-9B---\Cref{fig:supp_sequence_flow} tracks sequence-specific visual attention across both $\text{T2} \rightarrow \text{T1CE}$ and $\text{T1CE} \rightarrow \text{T2}$ directions under \emph{Think OFF}. During the shaded pre-injection stage, each model inspects the initial sequence free from competition by the complementary modality. Immediately following injection, both architectures actively engage the newly introduced sequence in either direction, revealing two distinct integration styles: Qwen3.5-9B shifts the majority of post-injection visual attention toward whichever sequence arrives second, whereas InternVL3.5-8B distributes post-injection attention more evenly across both acquisitions. Crucially, the strong post-injection surge onto $\text{T2}$ frames in Qwen3.5-9B under $\text{T1CE} \rightarrow \text{T2}$ argues against visual neglect as the sole driver of its directional asymmetry (\Cref{tab:primary_comparison} of the main paper): Qwen3.5-9B clearly attends to late $\text{T2}$ slices, consistent with conceptual anchoring to the initial hypothesis committed inside its longer $\text{T1CE}$ prefix cache (\Cref{sec:supp_cue_ablation}). Presenting $\text{T2}$ first prevents premature contrast-driven commitment, turning Qwen3.5-9B's strong post-injection focus on $\text{T1CE}$ into productive diagnostic refinement.

\subsection{Anatomy of Case-Level Prediction Revisions}
\label{sec:supp_flip_anatomy}

\Cref{fig:supp_flip_anatomy} connects these internal attention dynamics directly to directional case-level outcome transitions across all $90$ patients. Consistent with the marginal class-balance shifts in \Cref{sec:supp_class_balance}, appending a text-only \texttt{Wait} cue without new images does not encourage calibrated verification of cached visual evidence; instead, it triggers widespread, ungrounded label reversals---predominantly overturning valid MET predictions into false GBM calls---yielding negative net accuracy across all text-only runs in both models.

Replacing ungrounded continuation with staged $\text{T2} \rightarrow \text{T1CE}$ visual injection transforms revision behavior from noisy label churn into targeted diagnostic correction. Overall turnover drops substantially, and constructive $\text{Metastasis} \rightarrow \text{Glioblastoma}$ fixes---recovering GBM masked by $\text{T1CE}$ dominance under joint bundling---decisively outnumber regressions in both architectures. Architectural differences in prefix anchoring remain directly visible in the opposite $\text{T1CE} \rightarrow \text{T2}$ direction: InternVL3.5-8B achieves modest net refinement regardless of order, whereas Qwen3.5-9B trades corrections and regressions evenly when anchored on an initial $\text{T1CE}$ impression.

\subsection{Exploratory Slice-Level Attention}
\label{sec:supp_slice_audit}

Finally, we summarize the exploratory slice-level analyses referenced in \Cref{sec:res_mechanistic_audit} of the main paper, examining whether case-level diagnostic integration mirrors raw slice-level attention salience. Across normalized anatomical $\text{T2}$ slice depth (\Cref{fig:supp_boundary_sinks}), both backbones exhibit positional sinks on early boundary frames during the initial five generated tokens rather than uniform volumetric weighting. As decoding proceeds, Qwen3.5-9B relaxes this opening-slice bias and distributes attention more evenly across interior slices, whereas InternVL3.5-8B maintains a pronounced anchor on the first frame throughout generation.

Evaluating these raw slice weights against tumor-bearing slices in \Cref{fig:supp_slice_pitfalls}(a) highlights why static slice-level attention should not be conflated with diagnostic reasoning. Brief, temporally concentrated attention spikes fail to reliably isolate tumor-bearing slices across either model. Time-averaged slice attention achieves clear tumor-versus-normal separation in Qwen3.5-9B---consistent with its interior-slice engagement in \Cref{fig:supp_boundary_sinks}---while remaining near zero in InternVL3.5-8B due to its persistent boundary sink. Qualitatively, cross-architecture alignment with lesion-bearing slices emerges primarily when tracking step-wise relative slice prominence jointly with fluctuations in overall visual attention, rather than in static averages or isolated spikes. More importantly, stratifying cases by diagnostic correctness under $\text{T2} \rightarrow \text{T1CE}$ (\Cref{fig:supp_slice_pitfalls}(b)) reveals an informative dissociation: correctly diagnosed examinations do not exhibit higher raw tumor-slice attention than misclassified ones on either backbone, and InternVL3.5-8B realizes reliable staged diagnostic gains even without strong static slice separation.

To verify whether the underlying visual encoders capture spatially meaningful lesion structure independently of these decoding-time attention sinks, \Cref{fig:supp_feature_heterogeneity} examines mid-depth projected visual feature dispersion across $\text{T2}$ slice neighborhoods. For both GBM and MET, within-slice feature heterogeneity peaks on tumor-bearing slices (trending slightly higher on average in GBM, with overlapping confidence intervals) and tapers monotonically across adjacent ($\pm 1$), near ($\pm 2$), and distant ($>2$) normal slices. Taken together, these exploratory findings indicate that rich volumetric lesion cues are already encoded inside frozen VLM visual representations, while raw slice-attention salience reflects decoding-time positional heuristics rather than a causal determinant of diagnostic accuracy. Staged visual inference succeeds by organizing \emph{when} and \emph{in what order} complementary evidence enters the context, giving frozen models an uncluttered reasoning trajectory to integrate those cues across sequences.

\section{Ethical Considerations}
\label{sec:supp_ethics}

\paragraph{IRB Oversight and De-Identification.}
This study involved only retrospective analysis of historical clinical imaging records and publicly released benchmarks, with no prospective patient recruitment, clinical intervention, or direct patient interaction. Retrospective use of the institutional multi-center brain MRI cohort was reviewed and approved by the Institutional Review Boards (IRBs) and Ethics Committees of the participating academic hospitals. Prior to any computational evaluation, all DICOM headers, metadata, and protected health information were stripped during volumetric export (\Cref{sec:supp_preprocessing}), followed by manual verification to ensure that no personal identifiers remained in the derived 2D image files. The two external cohorts, PI-CAI~\cite{saha2023artificial} and OpenSwissHCC~\cite{tomography12090133}, were used strictly in accordance with their original public releases, de-identification standards, and governance terms.

\paragraph{Clinical Safety and Responsible Release.}
The VLMs evaluated in this work are studied solely as academic research probes and are not validated, cleared, or intended for clinical diagnosis, patient triage, or autonomous medical deployment. Upon acceptance, our de-identified evaluation data and inference code will be released publicly under research-only terms of use, accompanied by clear documentation of intended scientific scope and known limitations to prevent clinical or commercial misuse.

\end{document}